\documentclass[10pt,twocolumn,letterpaper]{article}

\usepackage[pagenumbers]{cvpr} % To force page numbers, e.g. for an arXiv version

\definecolor{cvprblue}{rgb}{0.21,0.49,0.74}
\usepackage[pagebackref=true,breaklinks=true,letterpaper=true,colorlinks,bookmarks=false,citecolor=cvprblue]{hyperref}

\usepackage{graphicx}
\usepackage{booktabs} % For formal tables
\usepackage{caption}
\usepackage{makecell}
\usepackage{multirow}
\usepackage{booktabs}
\usepackage{epstopdf}
\usepackage{marvosym}
\usepackage{amsmath}
\usepackage{colortbl}
\usepackage{changepage}
\usepackage{algorithm}
\usepackage{algorithmic}
\usepackage{amsmath}
\usepackage{amssymb}
\usepackage{tabularx}
\usepackage{bm}

\usepackage[accsupp]{axessibility}

\def\paperID{2725} % *** Enter the Paper ID here
\def\confName{CVPR}
\def\confYear{2026}

\makeatletter
\let\@oldmaketitle\@maketitle%
\renewcommand{\@maketitle}{\@oldmaketitle%
	\vskip -2em
	\centering
	\includegraphics[width=1\linewidth]{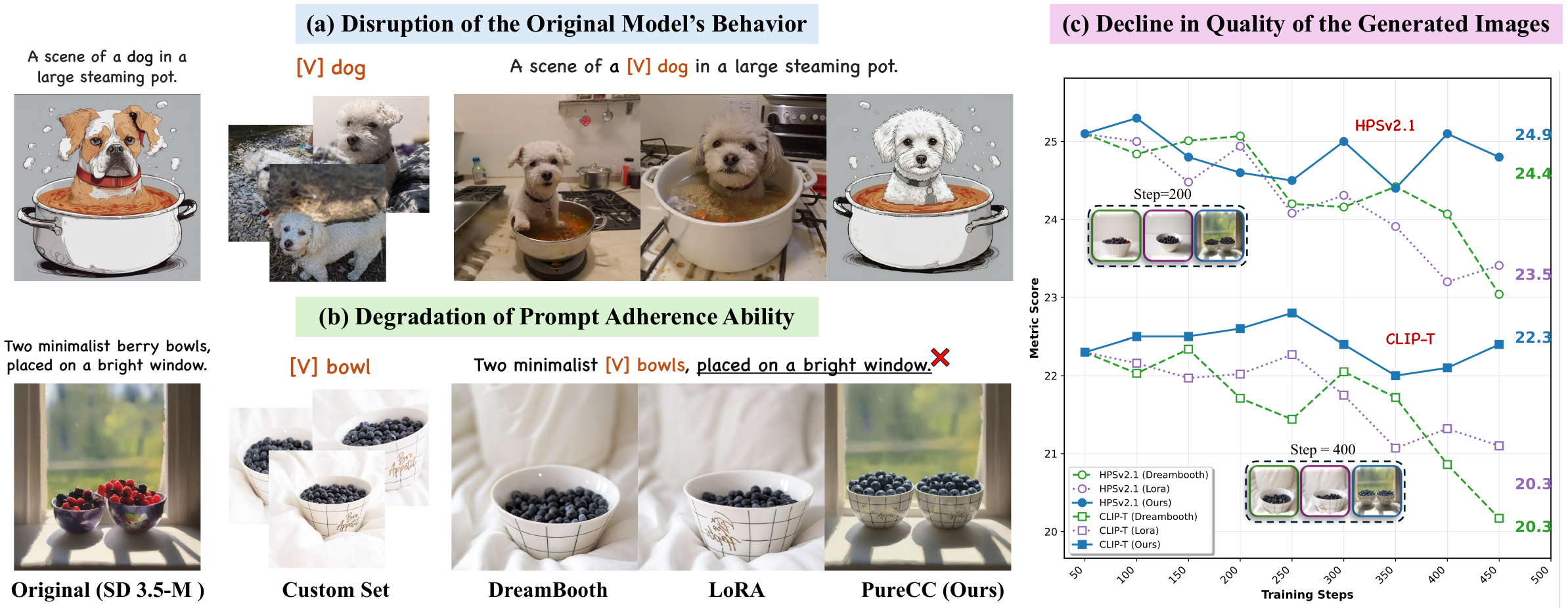}
	\vskip -0.5em
	\captionof{figure}{We introduce \textbf{\textit{PureCC}}, a novel concept customization approach. \textbf{(a)} PureCC effectively maintains target-unrelated image elements with original model's behavior after the personalized concept insertion. \textbf{(b)} Existing methods such as DreamBooth~\cite{ruiz2023dreambooth} and LoRA~\cite{hu2021lora} fail to follow the prompt `placed on a bright window' during custom generation. \textbf{(c)} The declined curve indicates that existing methods compromise the original model's ability of prompt adherence (CLIP-T~\cite{radford2021learning}) and generating high-quality images (HPSv2.1~\cite{wu2023human}).}
	\label{fig:teaser}
\bigskip}

\title{PureCC: \underline{Pure} Learning for Text-to-Image \underline{C}oncept \underline{C}ustomization }

\author{
    Zhichao Liao\textsuperscript{1\ * \dag}
    \enspace Xiaole Xian\textsuperscript{2\ * \dag}
    \enspace Qingyu Li\textsuperscript{4}
    \enspace Wenyu Qin\textsuperscript{4}
    \enspace Meng Wang\textsuperscript{4} 
    \enspace Weicheng Xie\textsuperscript{2,3 \Letter}
    \\
    \enspace Siyang Song\textsuperscript{5} 
    \enspace Pingfa Feng\textsuperscript{1}
    \enspace Long Zeng\textsuperscript{1\ \Letter}
    \enspace Liang Pan\textsuperscript{6}
    \vspace{0.33em}
    \\ \normalsize
    \textsuperscript{1}Tsinghua University \quad\enspace 
    \textsuperscript{2}School of Computer Science \& Software Engineering, Shenzhen University \quad\enspace \vspace{-0.25em}
    \\ \normalsize
    \textsuperscript{3}Guangdong Provincial Key Laboratory of Intelligent Information Processing, Shenzhen University \quad\enspace \vspace{-0.25em}
    \\ \normalsize
    \textsuperscript{4}Kling Team, Kuaishou Technology \quad\enspace
    \textsuperscript{5}University of Exeter \quad\enspace
    \textsuperscript{6}S-Lab, Nanyang Technological University
}

\begin{document}
\maketitle

{\let\thefootnote\relax\footnote{{\textsuperscript{*} Equal Contribution.}}}
{\let\thefootnote\relax\footnote{{\textsuperscript{\dag} This work was conducted during the author’s internship at Kling Team, Kuaishou Technology.}}}
{\let\thefootnote\relax\footnote{{\textsuperscript{\Letter} Corresponding author.}}}

\begin{abstract}
% \vspace{-0.5em}
% Concept customization has achieve remarkable outcomes in diverse application domains.
% However, recent methods primarily focus on high-fidelity and multi-concept customization, 

Existing concept customization methods have achieved remarkable outcomes in high-fidelity and multi-concept customization.
However, they often neglect the influence on the original model's behavior and capabilities when learning new personalized concepts.
To address this issue, we propose \textbf{\textit{PureCC}}. 
PureCC introduces a novel decoupled learning objective for concept customization, which combines the \textbf{implicit guidance} of the target concept with the \textbf{original conditional prediction}.
This separated form enables PureCC to substantially focus on the original model during training.
Moreover, based on this objective, PureCC designs a dual-branch training pipeline that includes a frozen extractor providing purified target concept representations as implicit guidance and a trainable flow model producing the original conditional prediction, jointly achieving pure learning for personalized concepts. 
Furthermore, PureCC introduces a novel adaptive guidance scale $\lambda^\star$ to dynamically adjust the guidance strength of the target concept, balancing customization fidelity and model preservation. 
Extensive experiments show that PureCC achieves state-of-the-art performance in preserving the original behavior and capabilities while enabling high-fidelity concept customization. 
The code is available at \url{https://github.com/lzc-sg/PureCC}.
\end{abstract}

\vspace{-1em}
\section{Introduction}
\label{sec:intro}

%%%%%%%%%%%%%%%%第一段%%%%%%%%%%%%%%%%%%
Concept customization~\cite{ruiz2023dreambooth, kumari2023multi}, an important task in custom text-to-image (T2I) generation, allows users to synthesize personalized concepts (e.g., new subjects or styles) contextualized in different scenes using only a few reference images (3-5).
Benefiting from the development of generative models like diffusion~\cite{rombach2022high, peebles2023scalable} and flow-based models~\cite{lipman2022flow, esser2024scaling}, it has attained impressive results in various application fields, including continuous content creation~\cite{chen2025omniinsert}, artistic production~\cite{frenkel2024implicit}, and advertising design~\cite{lin2025dreamfit,li2025anydressing}.

%%%%%%%%%%%%%%%%第二段%%%%%%%%%%%%%%%%%%
Most methods~\cite{ruiz2023dreambooth, gu2023mix, simsar2025loraclr} learn personalized concepts by adapting the distribution of pre-trained model to match the user-specific concept distribution through full fine-tuning or parameter-efficient techniques like LoRA~\cite{hu2021lora}. They generally associate the target concept with an identifier [V] and enable personalized generation via prompt injection during inference. However, existing research mainly emphasizes high-fidelity and multi-concept customization while overlooking two significant issues:

\noindent\textbf{$\bullet$ Disruption of the Original Model’s Behavior: }An ideal personalized concept insertion should focus solely on adjusting the concept-related aspects while keeping the image elements unrelated to the target concept consistent with the original model’s behavior.
However, as the case illustrated in Fig.~\ref{fig:teaser} (a), existing methods fail to alter only the original dog to the target [V] dog, with changing unrelated original image elements such as background, style and lighting.
This is because they treat all language-vision knowledge in the custom set as the learning source, but with limited reference images for learning, the generative model struggles to differentiate the target concept from other redundant information in the custom set, and to establish a unique association with the identifier [V].
Therefore, during custom generation, it leads to undesirable predictions of the target concept, disrupting the original model's behavior.
To our knowledge, such disruptions in concept customization have not been addressed or studied in previous works.

\begin{figure}[!t]
  \centering
  \includegraphics[width=1.0\linewidth]{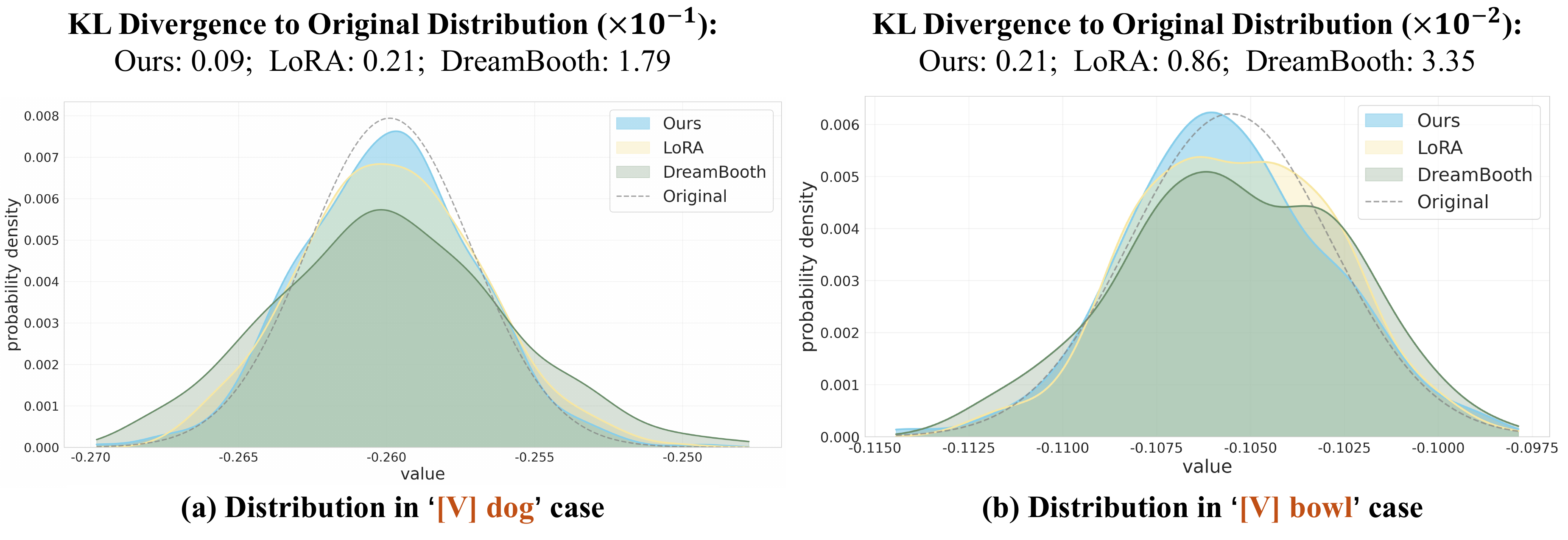}
  \vspace{-1.7em}
  \caption{\textbf{Original Distribution Drift. }Visualization and KL Divergence results demonstrated that existing methods, which adjust pre-trained models to align with the target distribution for learning personalized concepts, lead to distribution drift.}
  \vspace{-1.5em}
  \label{fig:distribution}
\end{figure}

\noindent\textbf{$\bullet$ Degradation of the Original Model’s Capability: }T2I generative models are pre-trained on large-scale multimodal databases, enabling them to effectively follow text prompts and generate high-quality images.
However, after transforming the pre-trained model into a custom model, existing methods tend to diminish these generative capabilities as shown in Fig.~\ref{fig:teaser} (b) and (c).
This issue arises because existing methods lack specific consideration for the original model in their learning objectives. Thus, when learning the personalized concept on scarce data, there is a risk of original data distribution drift as shown in Fig.~\ref{fig:distribution}, resulting in the degradation of the model's capability to adhere to prompt inputs and generate high-quality images.

To address these issues, in this paper, we propose \textit{\textbf{PureCC}}, a novel concept customization fine-tuning method that aims to purely learn personalized concepts while minimizing the influence on the original model's behavior and capability.
Specifically, PureCC designs an innovative learning objective to guide fine-tuning, which can be formed as a distinct combination of \textbf{implicit guidance of the target personalized concept} and \textbf{original conditional prediction}.
This separated form allows PureCC to practically consider the original model while learning the personalized concept.
To decouple the target concept from the custom set, we first introduce a representation extractor that uses a pre-trained flow-based model as the backbone, and we employ layer-wise tunable concept embeddings to fine-tune this model on the custom set, enhancing the understanding and representation of target concepts.
Then, we introduce our dual-branch training pipeline to purely learn the target concept while preserving original model's behavior and capability, which comprises the frozen representation extractor and a trainable flow model.  
During training, the frozen representation extractor provides a relatively pure representation of the target concept, while the trainable model offers a basic conditional prediction, serving as the implicit guidance and the original prediction for the proposed learning objective, respectively.
Moreover, we propose a novel adaptive guidance scale $\lambda^\star$ based on the representation alignment between dual branches to dynamically adjust the strength of the target concept guidance, effectively balancing the trade-off between personalized concept fidelity and original model preservation.
Extensive experiments show that our method achieves state-of-the-art performance in preserving the original behavior and capabilities while enabling high-fidelity concept customization. In summary, our contributions are as follows:

% We propose a new task named pure concept customization, which

% which requires generating personalized concepts while minimizing the impact on the original model. To achieve this goal, we introduce PureCC, a novel fine-tuning approach that reformulates an additional learning objective and designs a dual-branch training pipeline for pure learning of the target concept.
% To prevent disruption of the original behavior, we introduce a representation extractor and use pure target concept representations to implicitly guide concept learning.

\begin{itemize}

\item We introduce PureCC, a novel concept customization method, which reformulates a learning objective to purely learn the personalized concepts while minimizing the impact on the original model's behavior and capability. 

\item We design a dual-branch training pipeline based on our learning objective with a frozen representation extractor and a trainable model, providing specific implicit concept guidance and original conditional prediction.

\item We introduce an adaptive scale $\lambda^\star$ based on cross-branch representation alignment, dynamically balancing concept fidelity and preservation of original model.

\end{itemize}

\section{Related Work}
\label{sec:related}

% have achieved notable results in generating high-quality images.

\noindent\textbf{Diffusion and Flow-based Models} are recent mainstream generative models.
Diffusion models~\cite{ho2020denoising, song2020denoising, rombach2022high} aim to learn Stochastic Differential Equations (SDEs) that control the diffusion process.
Flow-based models~\cite{lipman2022flow} offer an alternative approach by directly modeling sample trajectories using Ordinary Differential Equations (ODEs) instead of SDEs.
Recent research~\cite{liu2022flow, esser2024scaling, zhuo2024lumina} has shown that ODE-based approaches attain faster convergence and improved controllability in T2I generation.
% Thus, flow-based models have become an attractive substitute for stochastic diffusion models.
In this study, we select the flow model SD3.5-M~\cite{esser2024scaling} as the basis for our research.

\noindent\textbf{Concept Customization} focuses on extending the pre-trained T2I model to generate personalized concepts. Existing methods can be categorized into Tuning-free methods~\cite{ye2023ip, xiao2025fastcomposer, zhang2024ssr, ma2024subject, dalva2025lorashop} and Tuning-based methods~\cite{ruiz2023dreambooth, kumari2023multi, hu2021lora, xian2025spf, dong2024continually, gu2023mix, gal2022image, guo2024pulid,huang2024classdiffusion}.
Tuning-free methods typically encode the reference image as feature embeddings and integrate them into the base models in a specific way (e.g., text embeddings~\cite{xiao2025fastcomposer} or the cross-attention layer~\cite{ye2023ip, zhang2024ssr}). 
Conversely, Tuning-based methods optimize specific parameters on a limited set of images to embed the personalized concept into the generative model.
DreamBooth~\cite{ruiz2023dreambooth} proposes to address subject-driven generation by fine-tuning all pre-trained model parameters, while several works employ textual inversion~\cite{gal2022image, voynov2023p+} to learn word embeddings of personalized concepts.
LoRA~\cite{hu2021lora} and its variants~\cite{dong2024continually, gu2023mix, zhong2024multi, lin2024lora, simsar2025loraclr} introduce additional low-rank subspaces to learn target concepts, reducing computational overhead.
Although existing methods have made significant progress in enhancing concept fidelity
and multi-concept customization, they often overlook the disruption of the original model's behavior and capabilities caused by concept insertion.

\noindent\textbf{Guidance in Generative Models} aims to achieve better controllability in the generation process~\cite{liao2024freehand}.
% One line of work is based on explicit guidance. 
Classifier Guidance~\cite{dhariwal2021diffusion} uses an additional pre-trained classifier to provide class-specific guidance for controllable generation and ~\cite{chefer2023attend, zhang2024enhancing, he2024aid, wang2024tokencompose, Cai_2025_CVPR} incorporate additional regularization guidance
% , such as semantic masks and segmentation maps, 
within the diffusion model to improve semantic perception and text-image alignment.
Although the above explicit guidelines are intuitive, they typically rely on external control components beyond the base model, making them less flexible and computationally demanding.
To address these issues, another line proposes implicit guidance. Classifier-Free Guidance (CFG)~\cite{ho2022classifier} treats the generative model itself as a conditional guidance branch, 
% and interpolates between conditional and unconditional outputs during sampling, 
thereby eliminating the need for an external classifier.
Subsequently, various studies~\cite{sadat2024eliminating, chung2024cfg++, cideron2024diversity, sadat2024no} have explored more advanced forms of implicit guidance to improve sample diversity and fidelity.
In this work, we extend the idea of implicit guidance into the training pipeline, formulating concept customization guided by implicit personalized concept representation, enabling pure and controllable concept insertion.

\section{Preliminary}
\label{sec:Preliminary}

\noindent\textbf{Conditional Flow Matching. }Suppose that $x_0 \sim q(x|y)$ is a data sample of the true distribution and $x_1 \sim p(x|y)$ represents a sample of source distribution. Recent conditional flow-based models adopt the Rectified Flow~\cite{liu2022flow} framework, which defines the source data sample $x_t$ as
\begin{equation}
\label{eq:flow_forward}
    x_t = (1-t)\,x_0 \;+\; t\,x_1,
\end{equation}
for \(t \in [0,1]\). Then a transformer model is trained to directly regress the velocity field $\frac{d}{dt}x_t = \bm{v}_t^\theta(x_t|y)$ by minimizing the Conditional Flow Matching (CFM) loss
\begin{equation}
\mathcal{L}_{CFM} = \mathbb{E}_{t, x_t} \left\| \bm{v}_t\big(x_t\big) - \bm{v}_t^\theta\big(x_t | y\big) \right\|_2^2,
\end{equation}
where the target velocity field is $\bm{v}_t\big(x_t\big) = x_1 - x_0$.

\noindent\textbf{Concept Customization. }Given a limited set of reference images with the same personalized concept, most methods optimize specific text tokens (e.g., [V]) to learn the target concept. 
In the custom set, the identifier [V] is typically combined with basic textual descriptions of the reference image (termed as Base text in this paper) to form a complete corpus.
In a case, the \textbf{Complete text} is `` \textit{A [V] dog standing on a surfboard, riding a wave}'', where ``\textit{A dog standing on a surfboard, riding a wave}'' is the \textbf{Base text} and ``\textit{[V] dog}'' is the \textbf{Target text}.
This Complete text is then encoded into textual embedding $y_{complete}$ by the pre-trained text encoder $E(\cdot)$ (e.g., CLIP~\cite{radford2021learning}, T5~\cite{raffel2020exploring}).
Finally, the pre-trained flow model achieves personalized concept learning by fine-tuning on a custom set using the loss:
\begin{equation}
\label{eq:concept_flow_loss}
    \mathcal{L}_{CC} = \mathbb{E}_{t, x_t} \left\| \bm{v}_t\big(x_t\big) - \bm{v}_t^\theta\big(x_t | y_{complete}\big) \right\|_2^2.
\end{equation}

% \bm{v}_t\big(x_t\big) - \bm{v}_t^\theta\big(x_t | y\big)
% where $y_{comb}$ is the text input including the concept embedding [v].
% Here, the flow matching model is trained by minimizing the loss \citep{lipman2022flow} expressed as: 
% the pretrained text encoder $E(\cdot)$ (e.g., CLIP \citep{radford2021learning}, T5 \citep{raffel2020exploring}) maps it to a textual embedding $y_{tar} = E(``[v]")$.
% During training, this token is typically used combined with other prompts as textual conditioning. 
% Custom diffusion models \citep{roich2022pivotal, gu2023mix, jiang2025mc, chen2024magic} (CDMs) utilize low-rank adaptation (LoRA) \citep{hu2021lora} to learn personalized concepts by finetuning the pretrained text-to-image diffusion model.

\noindent\textbf{Implicit Guidance. }In Classifier-Free Guidance (CFG) \cite{ho2022classifier}, a flow model $\bm{v}_t^\theta\big(x_t | y\big)$ is trained to predict both conditional and unconditional velocity fields.
This is achieved by introducing $y = \emptyset$, which denotes the null condition. 
During inference, the guided velocity field is formed by
\begin{equation}
\label{eq:CFG}
    \hat{\bm{v}}_t^\theta\big(x|y\big) = (1-w) \cdot \bm{v}^\theta_t\big(x|y=\emptyset\big) + w \cdot \bm{v}^\theta_t\big(x|y\big),
\end{equation}
where $w$ is the guidance scale. CFG treats the generative model itself as an implicit classifier to guide the generation process. To intuitively understand implicit conditional guidance, we can rewrite Eq.~\ref{eq:CFG} as follows:
\begin{equation}
\small
\label{eq:implicit-CFG}
    \hat{\bm{v}}_t^\theta(x|y) =  \bm{v}^\theta_t(x|y=\emptyset) + w \underbrace{ \cdot\big(\bm{v}^\theta_t\big(x|y) - \bm{v}^\theta_t\big(x|y=\emptyset)\big)}_{\textbf{Implicit Conditional Guidance}}
\end{equation}

% by introducing $y = \emptyset$,  which does not contain any conditional information.
% Classifier-Free Guidance (CFG) \cite{ho2022classifier} treats the generative model itself as an implicit classifier to guide the generation process and a single flow model $\bm{v}_t^\theta\big(x_t | y\big)$ is trained to output both conditional and unconditional velocity fields.

% In CFG, a single flow model $v_{t,\theta}(x_{t}|y)$ is trained to output both conditional and unconditional velocity fields. 
% This is done by introducing $y = \emptyset$, which does not contain any conditioning information.
% According to Bayes, the guidance gradient is computed via an implicit classifier $\nabla_x \log p(y\mid x)$, where $\nabla_x (\cdot)$ is the corresponding gradient and is used to steer the generation process during inference as:
% \begin{align}
% \begin{aligned}
% \label{eq:cfg}
% \Rightarrow \nabla_{\mathbf{x_{t}}}\log \hat{p}(\mathbf{x_{t}}\mid\mathbf{y}) &= \nabla_{\mathbf{x_{t}}}\log p(\mathbf{x_{t}}) + \\
% & \quad w \cdot \underbrace {\left(\nabla_{\mathbf{x_{t}}}\log p(\mathbf{x_{t}}\mid\mathbf{y}) - \nabla_{\mathbf{x_{t}}}\log p(\mathbf{x_{t}})\right)}_{\textbf{Implicit Guidance}}, \\
% \hat{\boldsymbol{v}}_{t,\theta}(x_t|y) = &(1-w) \cdot \boldsymbol{v}_{t,\theta}(x_t|y=\emptyset) + w \cdot \boldsymbol{v}_{t,\theta}(x_t|y),\\
% \end{aligned}  
% \end{align}
% where the $w$ is the guidance scale and $\emptyset$ represent the null-text input.

\section{Methodology}

\begin{figure*}[!t]
  \centering
  \includegraphics[width=1.0\linewidth]{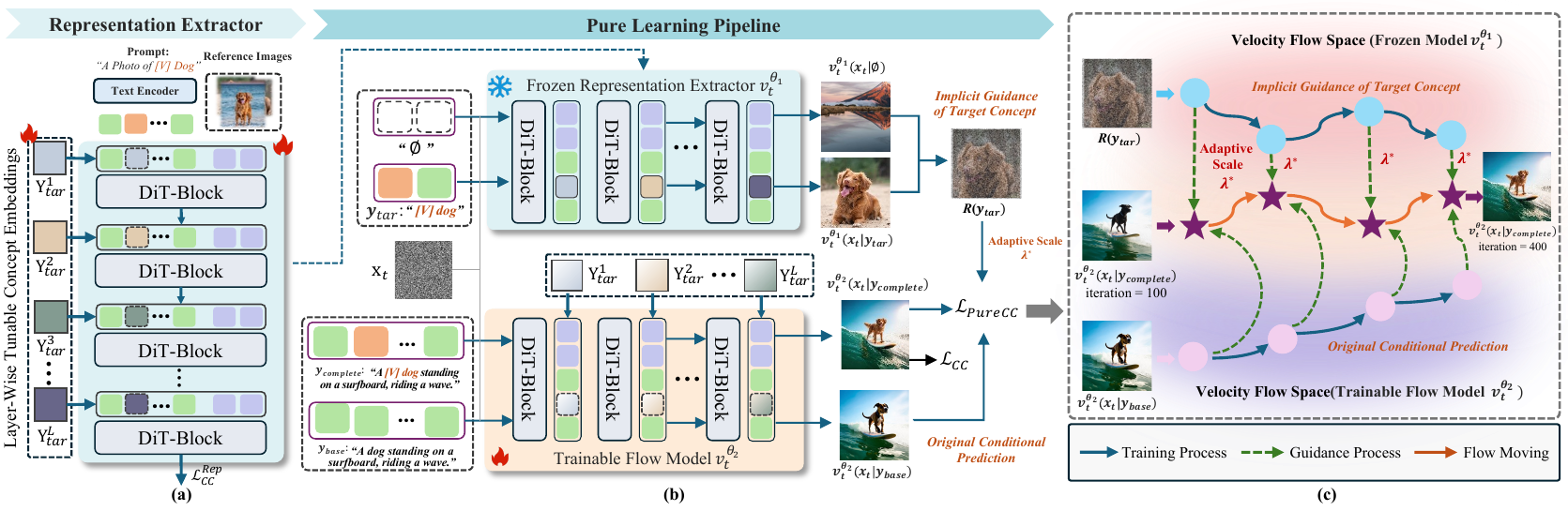}
  \vspace{-1.7em}
  \caption{\textbf{Overview of our PureCC.} \textbf{(a).} We first fine-tune a flow model on the custom set as representation extractor. \textbf{(b).} During the pure learning stage, the representation extractor remains frozen and provides the target concept representation, which is then controlled by our adaptive scale $\lambda^\star$ to \textbf{implicitly guide} the trainable model. The trainable model is initialized from another pre-trained flow model and provides \textbf{original conditional prediction }using the Base Text as input. The entire pipeline is trained on the custom set using $\mathcal{L}_{PureCC}$ and $\mathcal{L}_{CC}$. \textbf{(c).} demonstrates the process of using our designed $\mathcal{L}_{PureCC}$ to purely learn the target concept in the velocity flow space.}
  \label{fig:pipeline}
  \vspace{-1em}
\end{figure*}

\subsection{Learning Objective in PureCC}
\label{sec:Objective}

% In this work, we propose PureCC, a novel fine-tuning paradigm for purely generating custom concepts while minimizing the impact on the original model's behavior and capability.
% To address the limitations of existing methods, PureCC introduces a novel loss $\mathcal{L}_{PureCC}$ based on the traditional concept customization loss $\mathcal{L}_{CC}$ for PCC task.
% \begin{equation}
% \label{eq:Pcc_loss}
%     \mathcal{L}_{PCC} = \mathcal{L}_{CC} + \mathcal{L}_{PureCC}.
% \end{equation}

To address the limitations of existing methods, PureCC introduces a novel learning objective for PCC task.
Specifically, inspired by the form within CFG in Eq.~\ref{eq:implicit-CFG}, the guided velocity field of conditional generation can be viewed as adding an implicit conditional guidance to the unconditional prediction. 
Similarly, we consider the goal velocity field of concept customization as adding \textbf{implicit guidance of the target concept} to the \textbf{original conditional prediction}.
Therefore, we define the learning objective as a combination of the original model component and the target concept component:
\begin{equation}
\label{eq:Purecc_obejecive}
    {\bm{v}_t}^{PureCC} =  {\bm{v}}_t^{original}  + \lambda \cdot {\bm{v}}_t^{target},
\end{equation}
where $\lambda$ is a guidance scale like $w$ in Eq.~\ref{eq:implicit-CFG}.
This decoupled form enables the model to substantially focus on the original model while learning the target concept.

\subsection{Representation Extractor}
\label{sec:Extractor}
Existing methods use all custom language-vision data as a training source. However, due to the scarcity of reference images, they fail to decouple the target concept representation from the custom set as guidance for fine-tuning.
% , thereby creating a unique connection with the identifier [V] to achieve pure customization.
To alleviate this issue, PureCC first designs a representation extractor as shown in Fig.~\ref{fig:pipeline} (a), which treats a generative model as the backbone because of its powerful text-image understanding ability.
Specifically, we  employ LoRA~\cite{hu2021lora} to fine-tune a pre-trained flow model $\bm{v}^{\theta_1}_t(\cdot)$ on custom set. 

% based on the generative model for the personalized concept, as shown in Fig.~\ref{fig:pipeline}.
% 通过在不同层级引入概念标记，模型可以捕捉到目标概念的更多细微和详细的表示。这使得模型对概念的理解更加丰富和全面。

% can be formulated as:
\noindent\textbf{Layer-Wise Tunable Concept Embeddings. }To further enhance the model's understanding of personalized concept, we introduce layer-wise tunable concept embeddings $\{\mathbf{Y}_{tar}^{l}\}_{l=1}^{L}$ for each transformer layer, where $L$ denotes the total number of layers.
These tunable embeddings replace the original embedding of [V] in the input prompt embeddings. 
% after being processed by the text encoder. 
For example, the prompt “A photo of a [V] dog” is transformed into “A photo of a [$\text{V}^{l}$] dog” in each layer. Thus, the complete textural embedding for the $l$-th layer is:
\begin{equation}
\begin{aligned}
\label{eq:layer_embedding}
        \mathbf{Y}^{l}_{complete} = [y_{base}; \mathbf{Y}_{tar}^{l}].
\end{aligned}
\end{equation}
% Based on this textual representation and optimizing via the flow-matching objective in Eq.~\ref{eq:concept_flow_loss}, we obtain the naively fine-tuned model $v_{t, \theta_{1}}(\cdot)$, which serves as the implicit classifier for the concept [v] in the subsequent stage. 
Subsequently, the model is optimized using the loss in Eq.~\ref{eq:concept_flow_loss}, i.e.,
$\mathcal{L}^{Rep}_{CC} =\mathbb{E}_{t,x_0,x_1} \left\| (x_1 - x_0) - \bm{v}^{\theta_1}_t(x_t |y_{complete})\right\|_2^2$, where $y_{complete} = \{ \mathbf{Y}_{complete}^l\}^L_{l=1}$.
% denotes the complete textual embeddings obtained by concatenating the base-text embedding with the tunable concept embedding at each transformer layer.
By introducing tunable concept embeddings at different layers, the representation extractor can capture more detailed textures of the target concept, leading to a more comprehensive understanding. 
Notably, these embeddings will be preserved and used to replace the corresponding concept embeddings during the subsequent learning stage.

\subsection{Pure Learning Pipeline in PureCC}
We present our pure learning pipeline in Fig.~\ref{fig:pipeline} (b).
This pipeline utilizes a dual-branch architecture comprising: \textbf{(1)} a frozen representation extractor ${\bm{v}}_{t}^{\theta_1}(\cdot)$ sourced from Sec.~\ref{sec:Extractor}; \textbf{(2)} a trainable model initialized from another pre-trained flow model ${\bm{v}}_{t}^{\theta_2}(\cdot)$ to purely learn the target concept.

% To minimize disruption of the original behavior, 
\noindent\textbf{Implicit Guidance of the Target Concept. }In the frozen branch, to alleviate the disruption of the original behavior, the extractor endeavors to provide a relatively pure representation of the target concept, serving as implicit guidance for the trainable branch.
Specifically, based on our extractor's deep understanding of the target concept, we separately input the Target Text and the null condition ``$\emptyset$" into the extractor.  By subtracting their prediction outputs, we obtain the representation bias $\mathbf{R}(y_{tar})$, which contains abundant information related to the target concept, as our ${\bm{v}}_t^{target}$ in the learning objective:
\begin{align}
\label{eq:implicit_representation}
{\bm{v}}_t^{target} = \mathbf{R}(y_{tar}) = {\bm{v}}_{t}^{\theta_1}(x_t|y_{tar}) - {\bm{v}}_{t}^{\theta_1}(x_t|\emptyset),
\end{align}
where $y_{tar} = \{\mathbf{Y}_{tar}^l \}_{l=1}^L$ denotes the textural condition with the layer-wise target concept embeddings.

\noindent\textbf{Original Conditional Prediction. } In the trainable branch, the flow model takes an additional input $y_{base}$ to predict the corresponding velocity field ${\bm{v}}_{t}^{\theta_2}(x_t|y_{base})$.
Due to ${\bm{v}}_{t}^{\theta_2}(x_t|y_{base})$ sufficiently representing the performance of the original model, we employ it as ${\bm{v}}_t^{original}$ to substantially consider the original model:
\begin{equation}
\label{eq:Original}
\begin{aligned}
{\bm{v}}_t^{original} = {\bm{v}}_{t}^{\theta_2}(x_t|y_{base}).
\end{aligned}
\end{equation}

\begin{figure}[!t]
  \centering
  \includegraphics[width=1.0\linewidth]{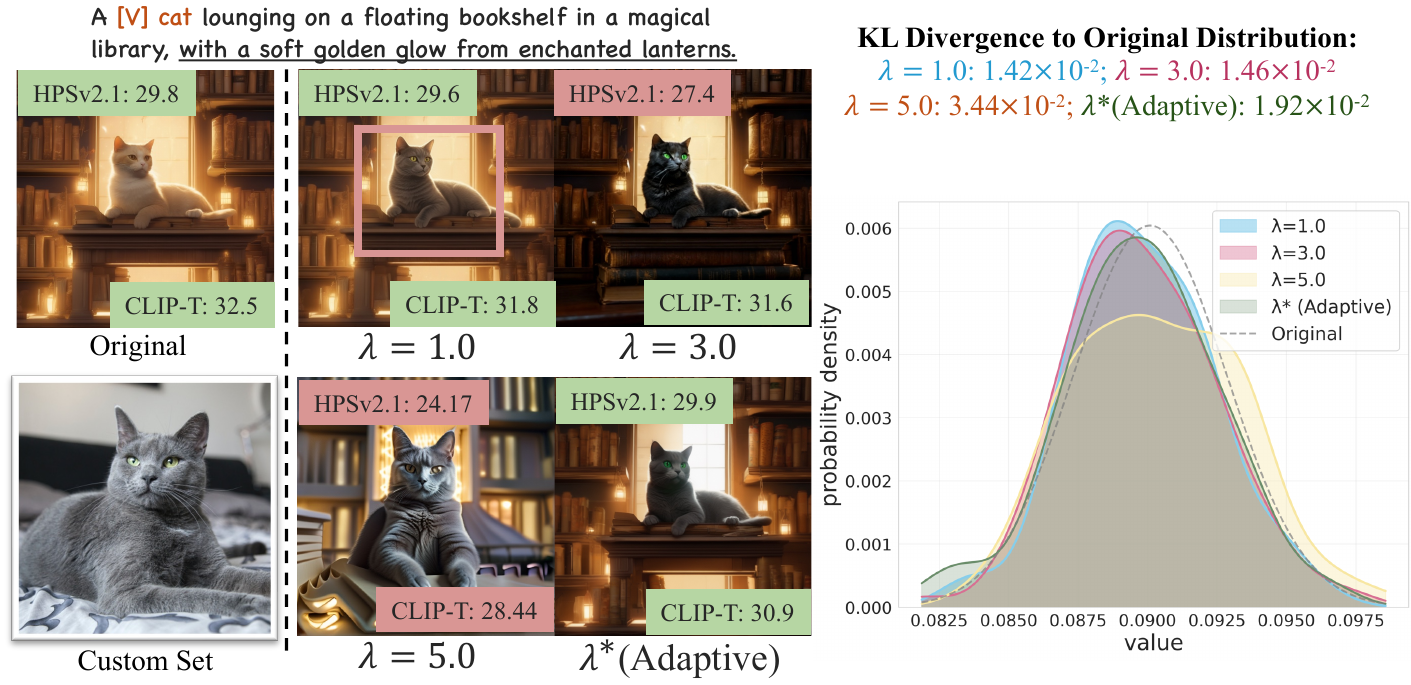}
  \vspace{-1.78em}
  \caption{\textbf{Motivation of Adaptive Scale $\lambda^\star$. }  A \textbf{small} $\lambda$ can \textbf{preserve the original model's behavior and capabilities} but leads to a \textbf{decrease in the fidelity} of the target concept. Conversely, when $\lambda$ is \textbf{excessively large}, the personalized concept dominates the learning objective, causing the final distribution to drift away from the original distribution. This results in a \textbf{degradation of the model’s generative ability}: the underlying prompt cannot be followed and lower CLIP-T and HPSv2.1 scores.}
  \label{fig:lambda}
  \vspace{-1.5em}
\end{figure}

% , which is essentially , represents the target concept through the  of the frozen model ${\bm{v}}_{t}^{\theta_1}(\cdot)$.
% Intuitively, $\lambda^\star$ serves as the projection coefficient of $\mathbf{R}(y_{complete}, y_{base})$ onto $\mathbf{R}(y_{tar})$
% from the frozen model ${\bm{v}}_{t}^{\theta_1}(\cdot)$, which  \
% Specifically, target concept representation in the frozen model ${\bm{v}}_{t}^{\theta_1}(\cdot)$ can be expressed as representation bias $R(y_{tar})$.

\subsection{Adaptive Guidance Scale $\lambda^\star$}
\label{sec:Adaptive}
Although the objective in Eq.~\ref{eq:Purecc_obejecive} is effective for purely learning personalized concepts, its performance relies on an unbounded empirical parameter $\lambda$, which controls the guidance strength of the target concept.
Improper tuning of this scale leads to undesirable artifacts, as illustrated in Fig.~\ref{fig:lambda}.
To balance this trade-off, we propose an adaptive mechanism to dynamically find the optimal $\lambda$. 
Specifically, the target concept representation used as the guidance can be expressed as the representation bias $R(y_{tar})$ of the frozen model ${\bm{v}}_{t}^{\theta_1}(\cdot)$.
Meanwhile, in the trainable model ${\bm{v}}_{t}^{\theta_2}(\cdot)$, we can acquire a similar form which represents the learned target concept representation:
\begin{equation}
\small
\begin{aligned}
\mathbf{R}(y_{complete}, y_{base}) = {\bm{v}}_{t}^{\theta_2}(x_{t}|y_{complete}) - {\bm{v}}_{t}^{\theta_2}(x_{t}|y_{base}).
\end{aligned}
\end{equation}
\noindent Then, we obtain the adaptive scale $\lambda^\star$ by minimizing the projection error between the learned representation $\mathbf{R}(y_{complete}, y_{base})$ in the trainable model and the guidance representation $\mathbf{R}(y_{tar})$ in the frozen model:
\begin{align} 
\lambda^\star = \arg\min_\lambda \|\mathbf{R}(y_{complete}, y_{base}) - \lambda \cdot \mathbf{R}(y_{tar})\|_2^2 
\end{align}
By differentiating, it can yield a closed-form solution:
\begin{align}
\label{eq:final_lambda}
\lambda^\star =  \frac{\langle \mathbf{R}(y_{complete}, y_{base}), \mathbf{R}(y_{tar}) \rangle}{\|\mathbf{R}(\mathbf{y}_{tar})\|^2},
\end{align}
where $<,>$ denotes the inner product of the corresponding representation. $\lambda^\star$ serves as the projection coefficient of $\mathbf{R}(y_{complete}, y_{base})$ on $\mathbf{R}(y_{tar})$
Intuitively, during training, if the trainable model has not yet learned the direction consistent with the guidance, $\lambda^\star$ will decrease to reduce the focus on the target concept and avoid contaminating the original model. Conversely, if the trainable model has learned the direction of the target concept relatively well, $\lambda^\star$ will increase to reinforce the learning of the target concept. This adaptive mechanism balances target concept fidelity and original model preservation.

\noindent\textbf{Overall Loss. }Our learning objective in Eq.~\ref{eq:Purecc_obejecive} is refined as:

{\footnotesize
\begin{equation}
\label{eq:PureCC_obj}
\begin{aligned}
{\bm{v}^{PureCC}_t} = \underbrace{{\bm{v}_t}^{\theta_2}(x_t|y_{base})}_{original} + \lambda^\star \underbrace{\cdot \big( {\bm{v}}_{t}^{\theta_1}(x_t|y_{tar}) - {\bm{v}}_{t}^{\theta_1}(x_t|\emptyset) \big)}_{target}.
\end{aligned}
\end{equation}}

\noindent And the training loss based on this objective is:
% 这样的学习目标，允许 trainable branch 学习纯净的目标概念知识 防止了 定制生成时 对原始行为的破坏 并 针对性的 关注了原始模型能力
% When using a custom set for pure learning, 
{\footnotesize
\begin{equation}
\label{eq:PureCC_loss}
    \mathcal{L}_{PureCC} = \mathbb{E}_{t, x_t} \left\| \bm{v}^{PureCC}_t - \bm{v}_t^{\theta_2}\big(x_t | y_{complete}\big) \right\|_2^2,
\end{equation}}

\noindent while Fig.~\ref{fig:pipeline} (c) illustrates the optimization process of this loss in the velocity flow space. As flow matching loss is responsible for predicting the velocity field and preserving the generative prior, we combine the loss in Eq.~\ref{eq:concept_flow_loss}, i.e., $\mathcal{L}_{CC} = \mathbb{E}_{t,x_0,x_1}\left\| (x_1 - x_0) - \bm{v}^{\theta_2}_{t}(x_t |y_{complete})\right\|_2^2$, with our proposed $\mathcal{L}_{PureCC}$. Therefore, the overall loss is:
\begin{align}
\label{eq:final_loss}
    \mathcal{L}_{PCC} = \mathcal{L}_{CC} + \eta \cdot \mathcal{L}_{PureCC},
\end{align}
where $\eta$ is the hyperparameter for regularization strength. Finally, as shown in Alg.~\ref{alg:purecc}, the training process enables PureCC to achieve pure learning for personalized concepts while minimizing the impact on the original model's behavior and capability.

\begin{algorithm}
\caption{PureCC Training Pipeline}
\begin{footnotesize} % 使用较小的字体
\begin{algorithmic}[1]
\REQUIRE Initialize flow model $v^{\theta_1}_{t}(\cdot)$; custom Set $\mathcal{D}$
\STATE \textbf{Training for Representation Extractor}
    \FOR{training iteration $k = 1$ \TO $K$}
    \STATE Sampling $(x, y_{complete})$ in $\mathcal{D}$
    \STATE Encode text prompts with layer-wise tunable embeddings: $\{\mathbf{Y}^l_{complete}\}^L_{l=1}$ in Eq.~\ref{eq:layer_embedding}
    \STATE Adapt the flow matching loss $\mathcal{L}^{Rep}_{CC}$ in Eq.~\ref{eq:concept_flow_loss} to optimize
    \STATE Update $\theta_1$ via LoRA fine-tuning
    \ENDFOR

\REQUIRE Initialize learnable model ${\bm{v}}_{t}^{\theta_2}(\cdot)$; Freeze the representation extractor ${\bm{v}}_{t}^{\theta_1}(\cdot)$; custom Set $\mathcal{D}$
\STATE \textbf{Pure Learning for Personalized Concept}
    \FOR{training iteration $k = 1$ \TO $K$}
    \STATE Sampling $(x, y_{tar}, y_{base}, y_{complete})$ in $\mathcal{D}$
    \STATE Compute implicit guidance of target concept  ${\bm{v}}_t^{target}=\mathbf{R}(y_{tar})$ in Eq.~\ref{eq:implicit_representation}
    \STATE Compute original conditional predictions and complete conditional predictions: ${\bm{v}}_t^{original} = {\bm{v}}^{\theta_2}_{t}(x_t|y_{base})$, and ${\bm{v}}^{\theta_2}_{t}(x_t|y_{complete})$
    \STATE Compute PureCC learning objective: $ {\bm{v}_t}^{PureCC} $ in Eq.~\ref{eq:PureCC_obj}
    \STATE Compute the adaptive scaling factor: $\lambda^\star$ in Eq.~\ref{eq:final_lambda}
    \STATE Compute the PureCC loss: $\mathcal{L}_{PureCC}$ in Eq.~\ref{eq:PureCC_loss}
    \STATE Optimize overall loss: $\mathcal{L}_{PCC}$ in Eq.~\ref{eq:final_loss}
    \STATE Update $\theta_2$
    \ENDFOR
    
\end{algorithmic}
\end{footnotesize} % 结束较小字体
\label{alg:purecc}
\end{algorithm}
\vspace{-1em}
\section{Experiments}
\label{sec:Experiment}

\begin{figure*}[!t]
  \centering
  \includegraphics[width=0.97\linewidth]{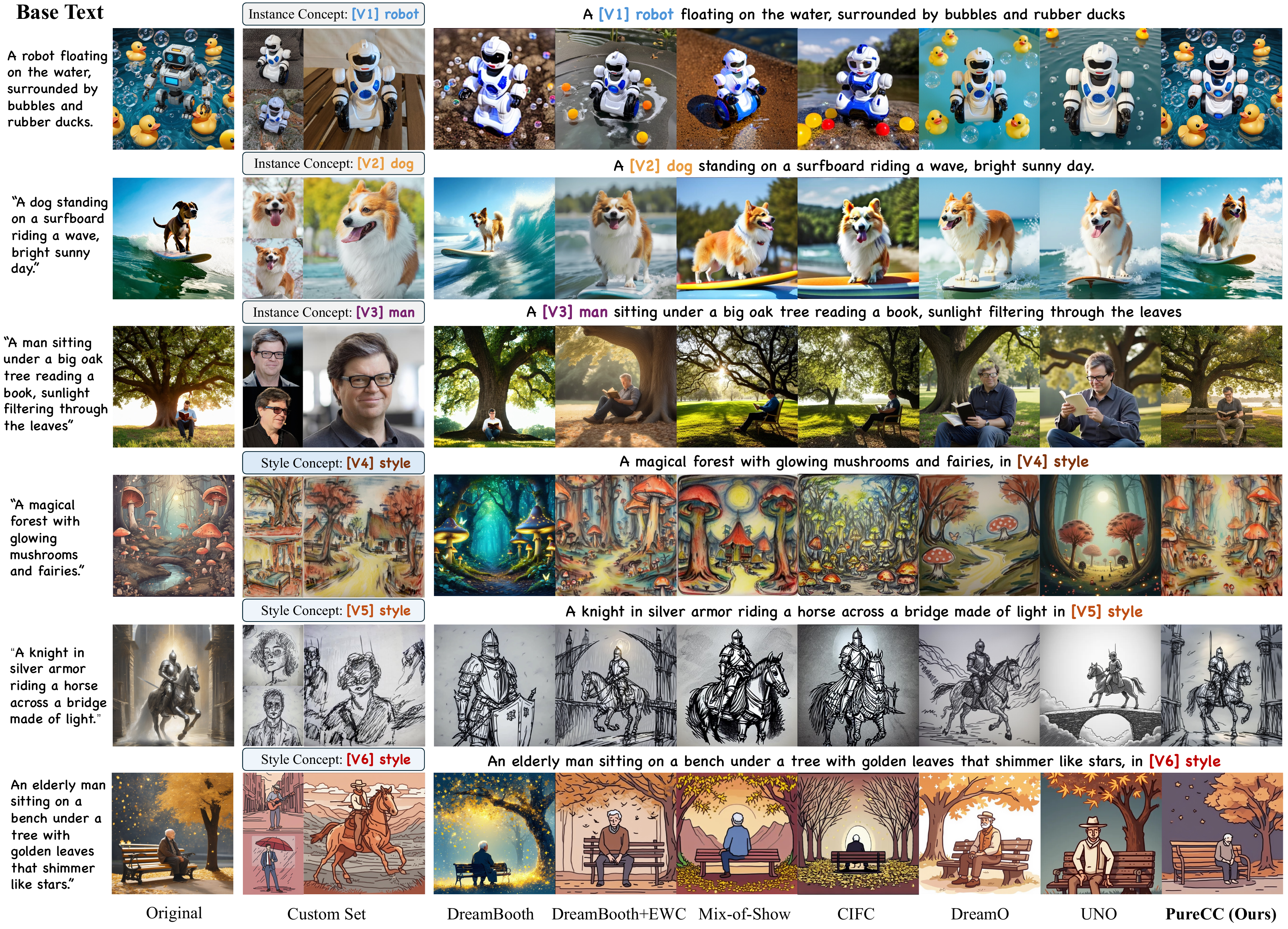}
  \vspace{-1em}
  \caption{\textbf{Qualitative Comparison with SOTAs} 
  % We compare our method with several recent customization approaches, 
  including Tuning-based methods: DreamBooth \citep{ruiz2023dreambooth}, DreamBooth + EWC \cite{serra2018overcoming}, Mix-of-Show \citep{gu2023mix}, CIFC \citep{dong2024continually}, and Tuning-free methods: DreamO \cite{mou2025dreamo}  UNO \cite{wu2025less}.
  }
  \label{fig:sota}
  \vspace{-1em}
\end{figure*}

\subsection{Experimental Setup}

% \noindent{\textbf{Implementation Details. }}
% In our experiments, to confirm the 

\noindent{\textbf{Dataset. }} 
To ensure a fair qualitative evaluation with previous methods, we select 14 personalized concepts from the dataset proposed by DreamBooth~\cite{ruiz2023dreambooth}.
Furthermore, to evaluate the adaptability of our method in broader scenarios, we additionally construct a batch of images containing 16 personalized concepts, covering both commonly used instance concepts (e.g., Pikachu, Yann LeCun) and style concepts (e.g., cartoon, sketch). 
% Part of the dataset is shown in the \textit{Appendix},and all data will be publicly released. 
For a comprehensive quantitative evaluation, we create DreamBenchPCC, which extends the DreamBench \cite{ruiz2023dreambooth} benchmark by adding an image set with 12 additional style concepts.
% Part of the data can be seen in the \textit{Appendix}.

% To ensure a fair comparison with previous methods, we adopt the same personalized datasets used in \cite{ruiz2023dreambooth}, which include 14 distinct personalized concepts.
% Furthermore, to evaluate the adaptability of our method in broader scenarios, we additionally construct a dataset containing 30 personalized concepts, covering both commonly used instance concepts (e.g., Pikachu, Yann LeCun) and style concepts (e.g., cartoon, sketch). Part of these datasets are shown in the \textit{Appendix}, and all collected samples will be publicly released.
% Therefore, for comprehensive evaluation, we create DreamBenchPCC, which extends the original DreamBench \cite{ruiz2023dreambooth} benchmark by adding a test set with 12 additional style concepts.
% 为了与先前的方法保持对比的公平，我们收集了他们所用的个性化数据集来自于\cite{a, b}，共14个个性化概念。并且为了验证我们方法在更广泛场景下的适配性，我们认为采集了一个包含30个个性化概念的数据集，包含一些网络上常见的instance概念（例如，皮卡丘，杨立坤）或是风格概念（例如，卡通，sketch）

\noindent{\textbf{Implementation Details.}} 
We adopt SD 3.5-M \cite{esser2024scaling} as our base model.
For a fair comparison, Tuning-based baselines such as DreamBooth~\cite{ruiz2023dreambooth}, B-LoRA~\cite{frenkel2024implicit}, LoRA-S~\cite{zhong2024multi}, and Mix-of-Show~\cite{gu2023mix} use the same pretrained backbone.
Following previous works~\cite{zhong2024multi, dong2024continually}, we set the LoRA rank to 4 and use a learning rate of $1.0\times10^{-4}$ to update both the flow model and the layer-wise tunable embeddings.

\noindent\textbf{Evaluation Metrics.}
For \textit{\textbf{target concept fidelity}}, we employ \textbf{CLIP-I (target)}~\cite{radford2021learning} and \textbf{DINO}~\cite{caron2021emerging} similarity to measure the consistency between generated images and the custom set for instance-level concepts, and \textbf{CSD}~\cite{somepalli2024measuring} (a CLIP-based style encoder) for style consistency.
% For the PureCC task, 
We introduce additional preservation metrics to quantify the custom model about the \textbf{\textit{preservation}} of the original model's \textbf{\textit{capability}}, including \textbf{$\Delta$CLIP-T (base)} for text alignment, \textbf{$\Delta$HPSv2.1} and \textbf{$\Delta$PickScore} for quality and aesthetic~\cite{liao2025humanaesexpert} preservation.
Where we report differential metrics $\Delta M = M_{\mathrm{custom}}(I(y_{\mathrm{complete}})) - M_{\mathrm{original}}(I(y_{\mathrm{base}}))$, $M$ denotes a base metric (e.g. CLIP-T~\cite{radford2021learning}, HPSv2.1~\cite{wu2023human}, PickScore~\cite{Kirstain2023PickaPicAO}) and $I(y_{\mathrm{base}})$ denotes generated image conditional on base text. The smaller $\Delta M$ indicates better preservation. 
\textbf{Seg-Cons}~\cite{kirillov2023segment} measures segmentation consistency between outputs of the custom model and the original model under the Complete text and Base text respectively, reflecting \textbf{\textit{behavior preservation}}.

\definecolor{tb-blue}{RGB}{196, 228, 252}

\begin{table*}[ht]
\centering
\caption{
\textbf{Quantitative Comparison Results on DreamBenchCC. } 
% $\Delta M$
% = M_{\mathrm{custom}}(I(y_{\mathrm{complete}})) - M_{\mathrm{original}}(I(y_{\mathrm{base}}))$, 
% where $M$ denotes a base metric, and $\Delta M$ measures the performance gap between the custom model and the original model (SD 3.5-M).
Since UNO and DreamO are Tuning-free methods that do not require fine-tuning the base model, our comparison for them focuses mainly on their concept responsiveness.
}
\label{tab:sota_comparsion}
\vspace{-1em}
\resizebox{1\textwidth}{!}{
\begin{tabular}{lcccccccccc}
\toprule
\multicolumn{1}{c}{\multirow{3}{*}{\textbf{Method}}} 
& \multicolumn{6}{c}{\textbf{Instance}} & \multicolumn{4}{c}{\textbf{Style}} \\
\cmidrule(r){2-7} \cmidrule(r){8-11}
& \multicolumn{4}{c}{\textbf{Preservation}} & \multicolumn{2}{c}{\textbf{Concept Responsiveness}} & \multicolumn{3}{c}{\textbf{Preservation}} & \multicolumn{1}{c}{\textbf{Concept Responsiveness}} \\
\cmidrule(r){2-5} \cmidrule(r){6-7} \cmidrule(r){8-10} \cmidrule(r){11-11}
& $\Delta$ CLIP-T (base)  ($\uparrow$) & $\Delta$ HPSv2.1 ($\uparrow$) & $\Delta$ PickScore ($\uparrow$) & Seg-Cons ($\uparrow$) & CLIP-I (target) ($\uparrow$) & DINO ($\uparrow$)  & $\Delta$ CLIP-T (base) ($\uparrow$) & $\Delta$ HPSv2.1 ($\uparrow$) & $\Delta$ PickScore ($\uparrow$) & CSD ($\uparrow$) \\ 
\midrule

\multicolumn{1}{l|}{Dreambooth \cite{ruiz2023dreambooth}}
& -4.81 & -2.17 & -3.90 & 18.38 & 0.63 & 0.62 & -6.23 & -2.08 & -1.83 & 0.57  \\
\multicolumn{1}{l|}{Dreambooth + EWC \cite{serra2018overcoming}}
& -4.17 & -2.20 & -3.16 & 26.37 & 0.62 & 0.61 & -7.90 & -2.04 & -1.57 & 0.60  \\
% \multicolumn{1}{l|}{LoRA-C \cite{zhong2024multi}}
% & -3.80 & -1.39 & -3.49 & 9.37 & 0.67  & 0.70 & -5.09 & -1.09 & -1.62 & 0.61  \\
\multicolumn{1}{l|}{Mix-of-Show \cite{gu2023mix}}
& -2.71 & -1.08 & -2.63 & 15.72 & 0.72  & 0.61 & -4.93 & -1.63 & -1.73 & 0.62 \\
\multicolumn{1}{l|}{CIFC \cite{dong2024continually}}
& -1.93 & -1.62 & -2.08 & 13.23 & 0.78 & 0.65 & -4.70 & -1.21 & -1.25 & 0.64  \\
\midrule
\multicolumn{1}{l|}{DreamO \cite{mou2025dreamo}}
& - & - & - & - &  0.71 & 0.67 & - & - & - & 0.65  \\
\multicolumn{1}{l|}{UNO \cite{wu2025less}} 
& - & - & - & - &  0.69 & 0.62 & - & - & -  & 0.34 \\
\midrule 
\rowcolor{tb-blue!40}
\multicolumn{1}{l|}{\textbf{Ours (PureCC)}} 
& \textbf{-0.31} & \textbf{+0.10} & \textbf{-0.67} & \textbf{69.37} & \textbf{0.81} & \textbf{0.73} & \textbf{-0.26} & \textbf{-0.92} & \textbf{-0.59} & \textbf{0.63} \\
\bottomrule
\end{tabular}
}
\end{table*}

% 我们的方法对比了一些最新的customization method，包括Mix-of-Show, LoRA-C, Dreambooth, CIFC.
% 我们也对比了UNO，DreamO这些tuning-free方法在概念响应上

\begin{figure}[!t]
  \centering
  \includegraphics[width=1.0\linewidth]{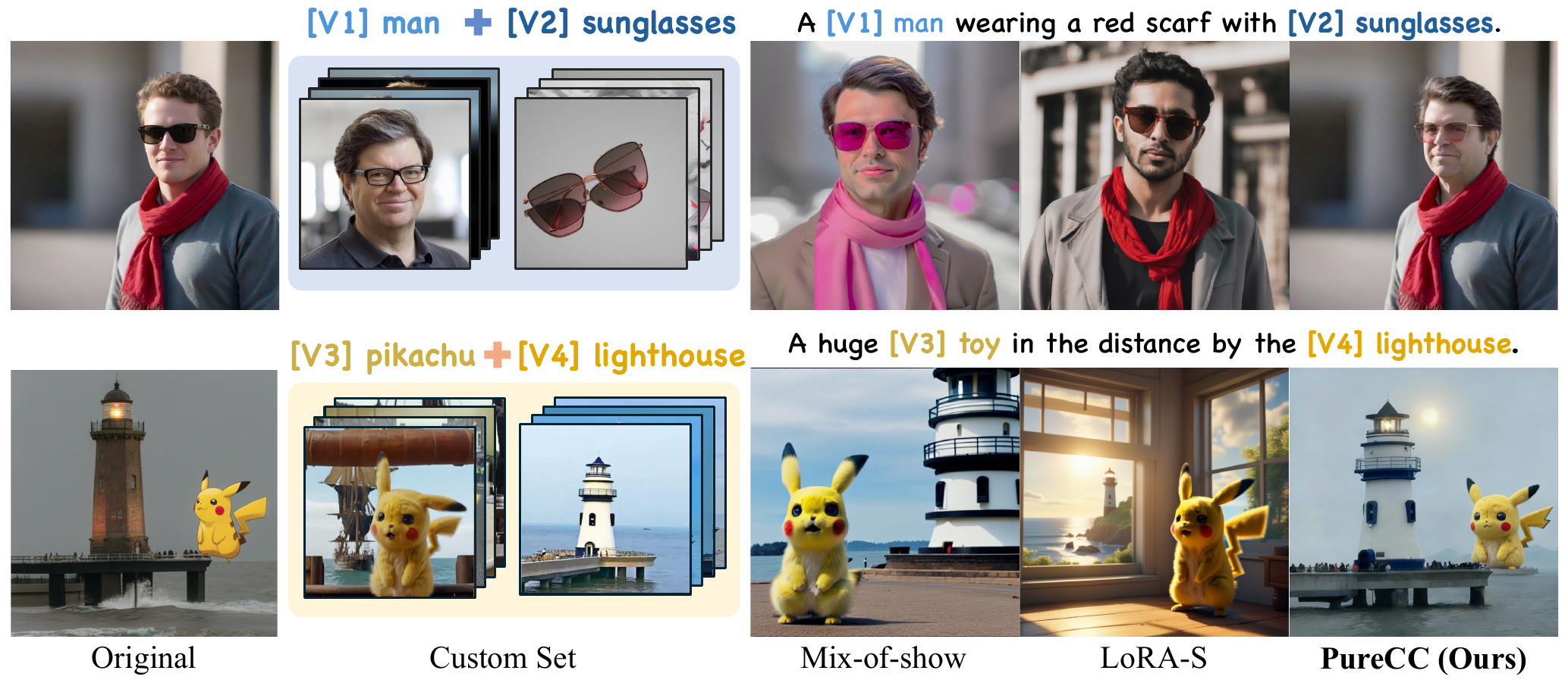}
  \vspace{-1.8em}
  % 不同方法的采样样本与原模型的分布差异
  \caption{
  \textbf{Qualitative comparison in Multi-Concept Customization} with Mix-of-show \cite{gu2023mix}, LoRA-S \cite{zhong2024multi}.}
  \label{fig:multi-concept}
  \vspace{-0.5em}
\end{figure}

% Qualitative comparison of Multi-Concept Customization.} Qualitative comparison of multi-concept composition results using different customization methods, including Mix-of-show \cite{gu2023mix}, LoRA-S \cite{zhong2024multi}. Each case involves combining two independently learned concepts.

\begin{figure}[!t]
  \centering
  \includegraphics[width=1\linewidth]{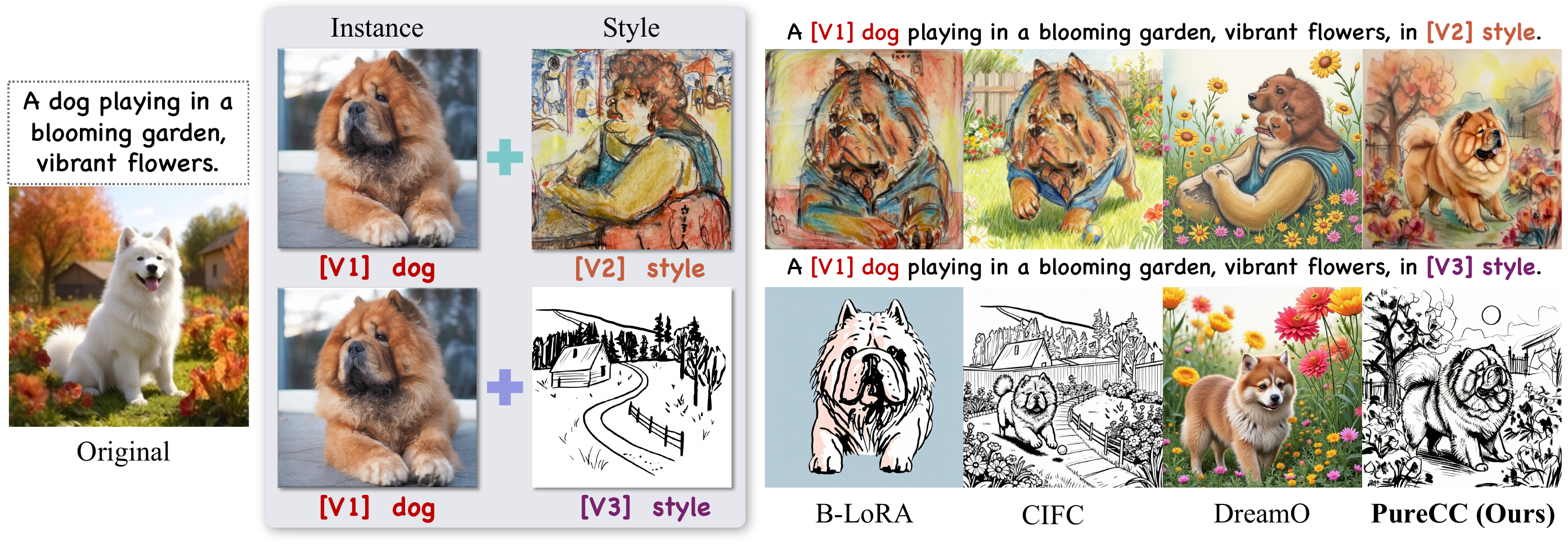}
  \vspace{-1.8em}
  \caption{\textbf{ Qualitative comparison of style–instance customization} across different methods, including CIFC \cite{dong2024continually}, B-LoRA \cite{frenkel2024implicit}, DreamO \cite{mou2025dreamo}. B-LoRA is a tuning-based approach specifically designed for balancing style and content adaptation. Each case combines an instance concept with a specific style.}
  \label{fig:stylized}
  \vspace{-1.5em}
\end{figure}

\subsection{Qualitative Evaluation}
\textbf{Single-Concept Customization.}
We compare our method with representative baselines
% , including DreamBooth \citep{ruiz2023dreambooth}, DreamBooth + EWC~\cite{serra2018overcoming}, Mix-of-Show \citep{gu2023mix}, and LoRA-C \citep{hu2021lora}, 
across both instance and style customization tasks. 
As illustrated in Fig.~\ref{fig:sota}, the baseline methods often fail to preserve the original model’s behavior and capabilities after learning new concepts. For instance, DreamBooth and Mix-of-Show exhibit severe inconsistency with the original behavior and alter global composition and background textures.
% \textcolor{red}{while LoRA-C struggles to maintain identity consistency in instance concepts such as 2-nd and 3-rd rows.} 
In contrast, our method achieves pure concept learning, which accurately adapts to new concepts while preserving non-target attributes such as background, lighting, and pose. These results show that our PureCC effectively produce high-fidelity customization without sacrificing the original model’s behavior and capabilities.

% mitigates capability degradation, producing high-fidelity customization without sacrificing the original model’s behavior.

\noindent{\textbf{Multi-Concept Customization.}}
% 我们方法CVA框架下对概念的纯净理解有利于多概念之间保持相对独立性而不相互污染，为了验证这一点，我们对比了tuing-based的方法在多概念个性化下效果。
% Theoretically, our 
PureCC encourages a disentangled and purified representation of each concept, allowing different customized concepts to remain relatively independent without semantic interference.
To validate this, we compare our method with several tuning-based approaches under multi-concept personalization settings.
As shown in Fig.~\ref{fig:multi-concept}, tuning-based methods such as Mix-of-Show and LoRA-S often suffer from semantic entanglement, where the adaptation of one concept unintentionally alters the appearance or context of another (e.g., color contamination between \textit{[V1] man} and \textit{[V2] sunglasses}, or structural distortion between \textit{[V3] pikachu} and \textit{[V4] lighthouse}.
In contrast, our method preserves the independence of each learned concept while integrating them coherently into a single composition.
This demonstrates that our PureCC enables pure multi-concept customization and effectively mitigates cross-concept interference.

% 为了验证我们方法风格、instance 这种不同概念组合场景下的效果，我们xxx

\noindent\textbf{Style-Instance Customization. } To further evaluate our capability in composing heterogeneous concepts such as instance and style, we conduct experiments on cross-domain customization scenarios.
As shown in Fig.~\ref{fig:stylized}, our method achieves a more balanced style transfer, faithfully preserving the object structure while accurately rendering the custom artistic style. In contrast, existing tuning-based approaches tend to overfit the style or distort object identity.

% demonstrating our method’s superior disentanglement and control.

\begin{figure*}[!t]
  \centering
  \includegraphics[width=0.87\linewidth]{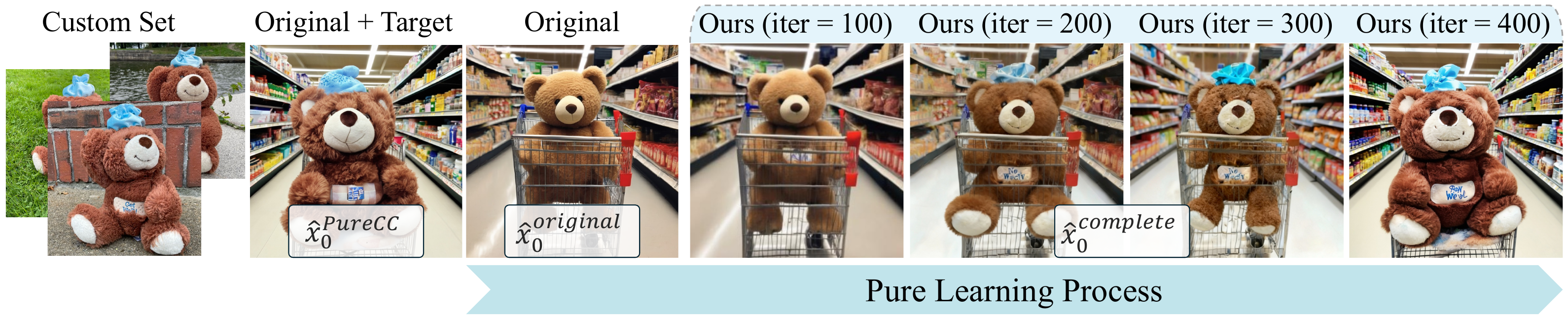}
  \vspace{-1em}
  \caption{
  \textbf{Visualization of Pure Learning Process. }$\widehat{x}_{0}^{PureCC}$ denotes the images obtained by integrating the velocity field $\{\bm{v}_{t}^{PureCC}\}_{t=1}^T$. Similarly, $\widehat{x}_{0}^{original}$ and $\widehat{x}_{0}^{complete}$ are based on $\{\bm{v}_{t}^{original}\}_{t=1}^T$ and $\{\bm{v}_{t}^{complete}\}_{t=1}^T$, respectively. ``iter" denotes the training iteration.
  }
  \vspace{-0.5em}
  \label{fig:process}
\end{figure*}

\noindent\textbf{Analysis of Predictions during Pure Learning.}
% 我们直观地展示了Pure Learning过程$\v^{\theta_2}_t(x_t\mid y_{complete})$的学习过程以及它的监督信号来xxx
% 为了研究
We intuitively visualize the Pure Learning process and its learning guidance.
% ${\bm{v}}_{t}^{\theta_2}(x_{t}|y_{complete})$ 
% ${\bm{v}}_t^{\mathrm{PureCC}}$.
% As shown in Fig.~\ref{fig:process}, the prediction $\widehat{x}_0^{\mathrm{complete}}$, obtained by integrating the velocity field，以$\widehat{x}_0^{\mathrm{original}}$为起点向$\widehat{x}_0^{\mathrm{PureCC}}$随着训练过程逐渐演变。
% 这表明在学习的过程中，模型incorporate the target semantic 而不影响 $y_{base}$所表达的内容。
As shown in Fig.~\ref{fig:process}, the prediction $\widehat{x}0^{\mathrm{complete}}$, evolves progressively during training: initially it is similar to $\widehat{x}_0^{\mathrm{original}}$ and gradually moves toward the the $\widehat{x}_0^{\mathrm{PureCC}}$ which both preserves the original model’s behavior and successfully expresses the target concept.
This process demonstrates that the objective $\mathcal{L}_{PureCC}$ purely incorporates the target concept while preserving the original content.
Overall, this visualization reveals that PureCC enables an additive and pure integration of new concepts, rather than disrupting the original model’s generative behavior.

\begin{table*}[t!]
\centering
\caption{
\textbf{Ablation Study on the Pure Learning.} ``Merged Learning Stage" refers to the training setting where the first-stage Representation Extractor $\bm{v}^{\theta_1}_{t}$ and the second-stage Pure Learning of $\bm{v}^{\theta_2}_{t}$ are conducted jointly.
}
\vspace{-0.5em}
\label{tab:abla}
\resizebox{0.87\linewidth}{!}
{
\begin{tabular}{lcccccc}
\toprule
\multicolumn{1}{c}{\multirow{2}{*}{\textbf{Strategy}}} 
& \multicolumn{4}{c}{\textbf{Preservation}} & \multicolumn{2}{c}{\textbf{Concept Responsiveness}} \\
\cmidrule(r){2-5} \cmidrule(r){6-7}
& $\Delta$ CLIP-T (base)  ($\uparrow$) & $\Delta$ HPSv2.1 ($\uparrow$) & $\Delta$ PickScore ($\uparrow$) & Seg-Cons ($\uparrow$) & CLIP-I (target) ($\uparrow$) & DINO ($\uparrow$) \\ 
\midrule

\multicolumn{1}{l|}{$\mathcal{L}_{CC}$}
&-4.52 & -2.01 & -2.95 & 23.74 & 0.65 & 0.66 \\
\multicolumn{1}{l|}{Merged Training Stage}
&-1.17 & -0.34 & -1.08 & 54.37 & 0.50 & 0.41 \\
\midrule \rowcolor{tb-blue!20}
\multicolumn{1}{l|}{$\mathcal{L}_{CC}$ +$\mathcal{L}_{PureCC}$ }
&\textbf{-0.31} & \textbf{+0.10} & \textbf{-0.67} & \textbf{69.37} & \textbf{0.81} & \textbf{0.73} \\

% \midrule 
% \rowcolor{tb-blue!40}
% \multicolumn{1}{l|}{\textbf{Ours}} &-0.31 & +0.10 & -0.67 & 69.37 & 0.81 & 0.73\\

\bottomrule
\end{tabular}
% \vspace{-2em}
}
\end{table*}

% \vspace{-2em}

\begin{figure}[!t]
  \centering
  \includegraphics[width=\linewidth]{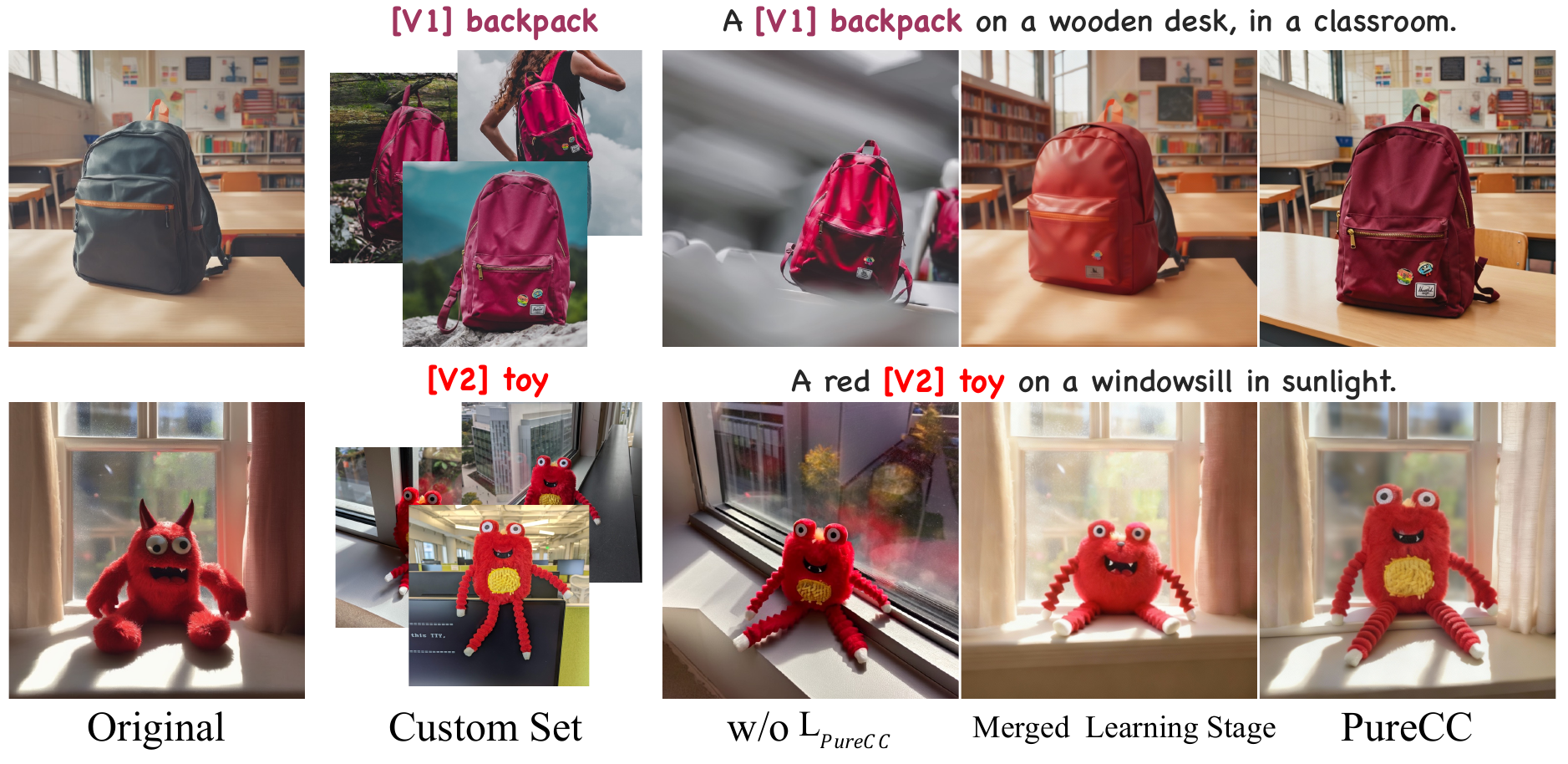}
  % \vspace{-1.5em}
  \captionsetup{skip=3pt}
  \caption{\textbf{Visualization of the Ablation Study.} 
  % ``Merged Learning Stage" refers to the training setting where the first-stage Representation Extractor $\bm{v}^{\theta_1}_{t}$ and the second-stage Pure Learning of $\bm{v}^{\theta_2}_{t}$ are conducted jointly.
  }
  \label{fig:abla}
  % \vspace{-1em}
\end{figure}

% \vspace{-1em}

\begin{table}[t]
\centering
\caption{
\textbf{Ablation Study of the $\lambda^\star$. }
}
\vspace{-0.5em}
\label{tab:abla2}
\resizebox{\linewidth}{!}
{
\begin{tabular}{lcccc}
\toprule
\multicolumn{1}{c}{\multirow{2}{*}{\textbf{Strategy}}}& \multicolumn{2}{c}{\textbf{Instance}} & \multicolumn{2}{c}{\textbf{Style}}\\
\cmidrule(r){2-3} \cmidrule(r){4-5}
  & $\Delta$CLIP-T (base) ($\uparrow$) & CLIP-I (target) ($\uparrow$) & $\Delta$CLIP-T (base) ($\uparrow$) & CSD ($\uparrow$) \\
\midrule
$\lambda = 1.0$ & -0.18 & 0.43 & -0.67 & 0.26\\
$\lambda = 3.0$ & -0.51 & 0.58 & -0.93 & 0.61\\
$\lambda = 5.0$ & -2.67 & 0.73 & -4.21 & 0.42\\
\midrule
\rowcolor{tb-blue!20}
$\lambda = \lambda^{\star}$ (adaptive) & \textbf{-0.31}  & \textbf{0.81} & \textbf{-0.26} & \textbf{0.63} \\
\bottomrule
\end{tabular}
}
\end{table}

% \vspace{-2em}

\subsection{Quantitative Evaluation}
Tab.~\ref{tab:sota_comparsion} presents the quantitative comparison of our method and existing approaches on the DreamBenchPCC.
Across both instance and style concept customization, our method consistently achieves superior performance in \textbf{Preservation} and \textbf{Concept Responsiveness} metrics.
In the Preservation aspect, our method attains the smallest gaps in $\Delta$CLIP-T(base), $\Delta$HPSv2.1, and $\Delta$PickScore, indicating that it best preserves the original model’s capability, including semantic alignment, aesthetic quality, and human preference.
Furthermore, the high Seg-Cons score (69.37) demonstrates that our approach maintains the spatial and structural consistency of the original model’s outputs, effectively mitigating behavioral disruption commonly observed in tuning-based methods such as DreamBooth and CIFC.
In terms of Concept Responsiveness, our approach achieves competitive or superior scores on CLIP-I(target), DINO, and CSD, suggesting that we can accurately express both instance and style personalized concepts without compromising generative fidelity.
These results validate that PureCC effectively integrates new concepts in a stable and semantically aligned manner, achieving state-of-the-art overall performance.

\subsection{Ablation Study}

% showcase the quantitative and qualitative results from this study.
% Compared with the baseline that optimizes only $\mathcal{L}_{CC}$, incorporating the PureCC loss ($\mathcal{L}_{CC} + \mathcal{L}_{PureCC}$) significantly improves all preservation metrics—including $\Delta$CLIP-T(base), $\Delta$HPSv2.1, and $\Delta$PickScore—while maintaining strong concept responsiveness in CLIP-I(target) and DINO.

\noindent\textbf{Pure Learning. }To confirm the importance of both the loss design and the two-stage training strategy in our PureCC, we perform quantitative and qualitative ablation studies in Tab.~\ref{tab:abla} and Fig.~\ref{fig:abla} 
Compared to the baseline, which optimizes solely $\mathcal{L}_{CC}$, the integration of the PureCC loss ($\mathcal{L}_{CC} + \mathcal{L}_{PureCC}$) leads to substantial enhancements in all preservation metrics including $\Delta$CLIP-T(base), $\Delta$HPSv2.1, and $\Delta$PickScore, while maintaining strong concept responsiveness in CLIP-I(target) and DINO.
These results validate that the PureCC objective effectively prevents the degradation of prior knowledge during fine-tuning, thereby preserving both the behavior and capability.
In the "Merged Learning Stage" setting, although joint optimization preserves the original model, the representation extractor does not adequately learn the target concept representation. As a result, the guidance becomes under-expressive, leading to a significant decline in the fidelity of the target concept.

\noindent\textbf{Adaptive $\lambda^{\star}$.} 
%量化的实验结果进一步验证了图4所表现的问题，说明了我们自适应scaler的必要性。如表Tab.~\ref{tab:abla2}所示，$\lambda=1$过小的情况下，纯净概念的引导有限，xxx
The quantitative results further validate the issues shown in Fig.~\ref{fig:lambda}, demonstrating the necessity of our adaptive scale. As shown in Tab.~\ref{tab:abla2}, using a fixed $\lambda$ leads to clear limitations. When $\lambda = 1$ is too small, the concept guidance becomes insufficient, resulting in weak adaptation to the target concept with noticeably lower CLIP-I (0.43) and CSD (0.26) scores. Increasing $\lambda$ strengthens concept responsiveness, but simultaneously harms preservation, as reflected by larger degradation in $\Delta$CLIP-T(base).

% \vspace{-1em}

% \vspace{+1em}
\section{Conclusion}
PureCC effectively addresses the challenge of preserving the original model's behavior and capabilities while achieving high-fidelity concept customization.
By introducing a decoupled learning objective and a dual-branch training pipeline, PureCC ensures pure learning for personalized concepts.
The adaptive guidance scale $\lambda^\star$
further enhances the balance between customization fidelity and model preservation. Extensive experiments demonstrate that PureCC outperforms existing methods in maintaining the original model while enabling concept customization.

% In this paper, we presented \textbf{Compositional Velocity Alignment (CVA)}, a novel framework for pure concept customization in diffusion models.
% Unlike conventional fine-tuning or PEFT-based approaches that often suffer from semantic drift and overfitting on few-shot data, our method reformulates customization as an \emph{additive velocity composition} guided by implicit concept gradients.
% Through a dual-branch design—comprising a frozen reference branch and a learnable response branch—CVA enables precise concept injection while preserving the base model’s generative prior.
% Furthermore, the adaptive scaling mechanism dynamically regulates the strength of gradient-based concept integration, effectively mitigating distribution collapse and overfitting.
% Extensive experiments demonstrate that CVA achieves superior semantic fidelity, visual quality, and generalization compared to state-of-the-art methods.
% While CVA demonstrates strong performance in few-shot personalization, it still relies on accurate gradient estimation from the reference branch, which can be sensitive to noisy or ambiguous target concepts.
% Future work will explore more efficient gradient modulation strategies and cross-concept regularization to further enhance scalability and stability.

\newpage
\section{Acknowledgement}
This work was supported by the National Key Research and Development Program of China (Grant No. 2022YFB3303101), the National Natural Science Foundation of China (Grant No. 62276170), and the Guangdong Provincial Key Laboratory (Grant No. 2023B1212060076).

% \clearpage
{
    \small
    \bibliographystyle{ieeenat_fullname}
    \bibliography{main}
}

\clearpage
\maketitlesupplementary

% {
% \small
% \tableofcontents
% }

% \vspace{35em}

\section{Dataset Details}
\begin{figure*}[!t]
  \centering
  \includegraphics[width=\linewidth]{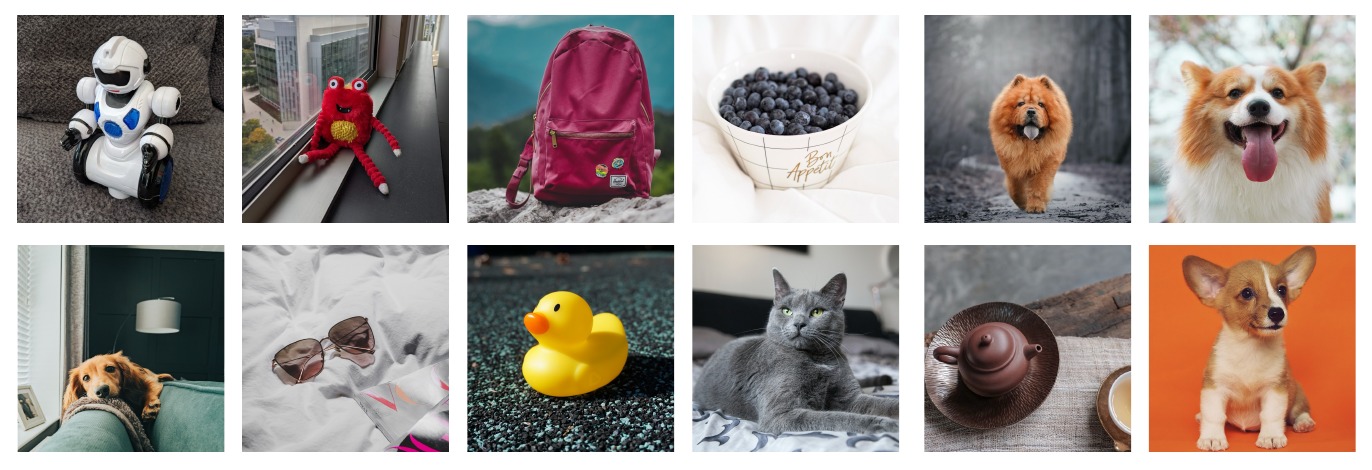}
  \caption{\textbf{Some samples selected from the dataset proposed by DreamBooth}~\cite{ruiz2023dreambooth}.}
  \label{fig:data1}
\end{figure*}

\begin{figure*}[!t]
  \centering
  \includegraphics[width=\linewidth]{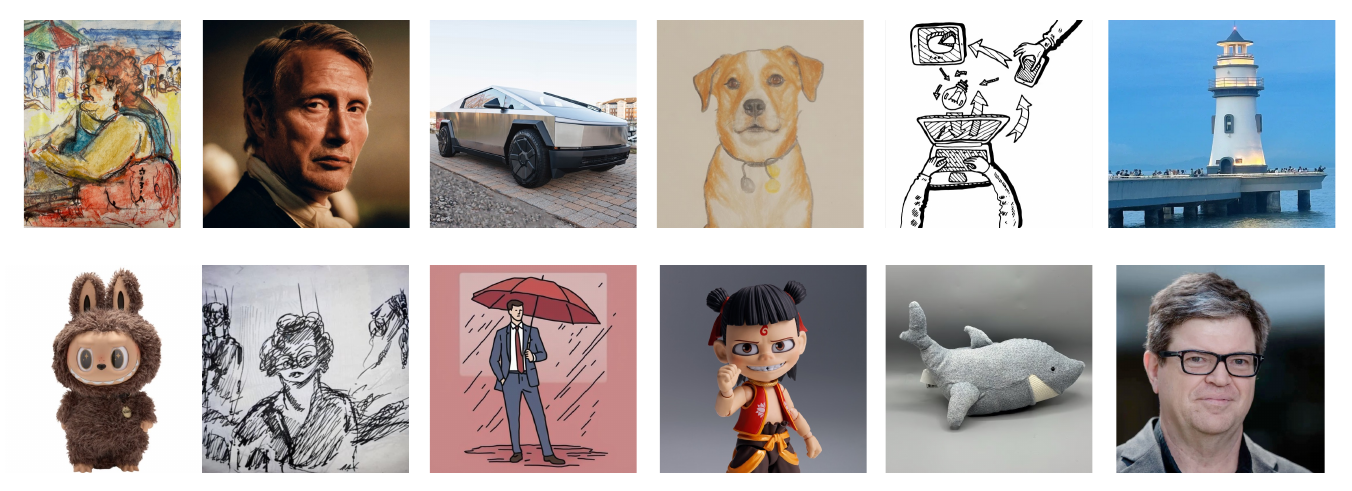}
  \caption{\textbf{Some samples additionally collected by us.}}
  \label{fig:data2}
\end{figure*}

\begin{figure*}[!t]
  \centering
  \includegraphics[width=\linewidth]{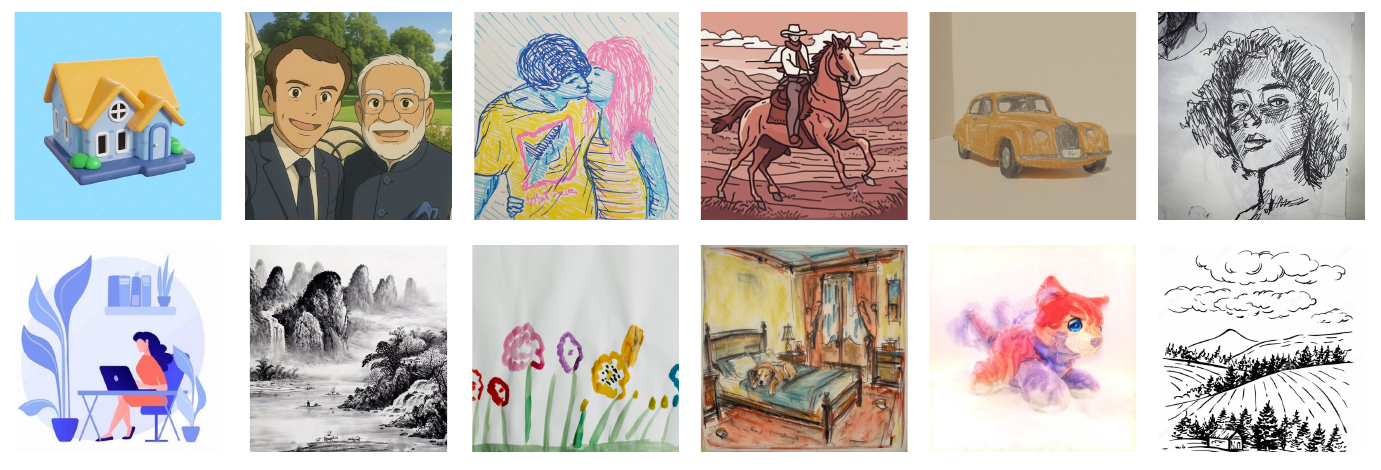}
  \caption{\textbf{Some style samples in our DreamBenchPCC.}}
  \label{fig:data3}
\end{figure*}

To ensure a fair \textbf{Qualitative Evaluation} with previous methods, we selected 14 personalized concepts from the dataset proposed by DreamBooth~\cite{ruiz2023dreambooth}. Some samples can be seen in Fig.~\ref{fig:data1}. Furthermore, to assess the adaptability of our method across a wider range of scenarios, we additionally collected a batch of novel personalized concepts, which includes 11 commonly used instance concepts, such as Pikachu and Yann LeCun, as well as 5 style concepts, such as cartoon and sketch. Some samples can be seen in Fig.~\ref{fig:data2}. Thus, we constructed a Qualitative Evaluation dataset comprising a total of 30 personalized concepts.
For comprehensive \textbf{Quantitative Evaluation}, we created DreamBenchPCC, which extends DreamBench~\cite{ruiz2023dreambooth} with 12 additional style concepts to balance the proportion of instance and style concepts.
Some style samples can be seen in Fig.~\ref{fig:data3}.
We used the state-of-the-art large multi-modal model Claude 3.5 Sonnet~\cite{Claude} to caption all newly collected images.

\section{More Implementation Details}
We perform training on an NVIDIA A100 GPU with a batch size of 2.
For each personalized concept, both the representation extractor $\bm{v}^{\theta_1}_t$ and the trainable model $\bm{v}^{\theta_2}_t$ are trained in 400 steps.
All images are generated using the default inference setting of 28 timesteps.

% 我们使用A100 进行训练，设定batchsize为2。对于每个属性，我们的Representation Extractor$\bm{v}^{\theta_1}_t$以及Pure customization model $\bm{v}^{\theta_2}_t$均进行400个steps的训练。所有图片维持默认的28个timestep推理步骤

% \clearpage

\section{Evaluation Metrics Details}
Since we are working with a new task setting—Pure Concept Customization—we specifically applied representative metrics to suit our task setting for quantitative evaluation.

\noindent \textbf{Fidelity of the personalized concept. }For instance-level concepts, we employ \textbf{CLIP-I (target)}~\cite{radford2021learning} and \textbf{DINO}~\cite{caron2021emerging} to evaluate the similarity between the target concept in the generated images and the target concept in the reference images from the custom set. For style-level concepts, we use \textbf{CSD}~\cite{somepalli2024measuring} (a CLIP-based style encoder) to evaluate style consistency.

\noindent \textbf{Original model preservation. }To evaluate the custom model's preservation of the original model's capabilities,  we report differential metrics $\Delta M = M_{\mathrm{custom}}(I(y_{\mathrm{complete}})) - M_{\mathrm{original}}(I(y_{\mathrm{base}}))$, where $M$ denotes a base metric (e.g. CLIP-T~\cite{radford2021learning}, HPSv2.1~\cite{wu2023human}, PickScore~\cite{Kirstain2023PickaPicAO}).
$M_{\mathrm{custom}}(I(y_{\mathrm{complete}}))$ represents the metric score of the image generated using the custom model and the Complete text, and $M_{\mathrm{original}}(I(y_{\mathrm{base}}))$ represents the metric score of the image generated using the original model and the Base text.
Thus, the smaller $\Delta M$ indicates better preservation.
We specifically use \textbf{$\Delta$CLIP-T(base)} to assess the ability to follow Base text, as well as \textbf{$\Delta$HPSv2.1} and \textbf{$\Delta$PickScore} to evaluate the retention of the ability to generate high-quality and aesthetically pleasing images.
Moreover, we use \textbf{Seg-Cons}~\cite{kirillov2023segment} to measures segmentation consistency between outputs of the custom model and the original model under the Complete text and Base text respectively, reflecting original behavior preservation.

% $M_{\mathrm{custom}}(I(y_{\mathrm{complete}}))$ 代表 使用custom model 和 Complete text 来生成的 图片的 指标score and $M_{\mathrm{original}}(I(y_{\mathrm{base}}))$代表使用原始模型 和 Base text 来生成的 图片的 指标 score
% we specifically use \textbf{$\Delta$CLIP-T(base)} to assess the ability to follow Base text, as well as \textbf{$\Delta$HPSv2.1} and \textbf{$\Delta$PickScore} to evaluate the retention of the ability to generate high-quality and aesthetically pleasing images.
% $I(y_{\mathrm{base}})$ and denotes generated image conditional on Base text. The smaller $\Delta M$ indicates better preservation. 

\section{Qualitative and Quantitative Evaluation Details.}
We performed qualitative evaluations on our personalized dataset, which has been expanded to include new instance and style concepts. This dataset comprises a total of 30 personalized concepts, as shown in
Fig.~\ref{fig:data1} and Fig.~\ref{fig:data2}.
For a comprehensive quantitative evaluation, we utilized DreamBenchPCC, as shown in Fig.~\ref{fig:data3}
For a fair comparison, tuning-based baselines such as DreamBooth~\cite{ruiz2023dreambooth}, B-LoRA~\cite{frenkel2024implicit}, LoRA-S~\cite{zhong2024multi}, and Mix-of-Show~\cite{gu2023mix} are all trained using the same pretrained backbone, SD 3.5-M~\cite{esser2024scaling}.
For tuning-free baselines such as DreamO~\cite{mou2025dreamo} and UNO~\cite{wu2025less}, we follow their standard usage and provide one reference image from the customization set with prompts to enable personalized generation.
For stylization cases, we prepend the prompts with the instruction “Generate a same-style image” to ensure consistent style conditioning.
Since tuning-free methods do not require fine-tuning the pre-trained model, they fundamentally differ from our task setting, which addresses the damage to the original model caused by adapting the pre-trained model to learn personalized concepts during fine-tuning.
Therefore, when comparing with them, we primarily focus on evaluating their outstanding performance in target concept fidelity.

\begin{figure*}[!t]
  \centering
  \includegraphics[width=0.76\linewidth]{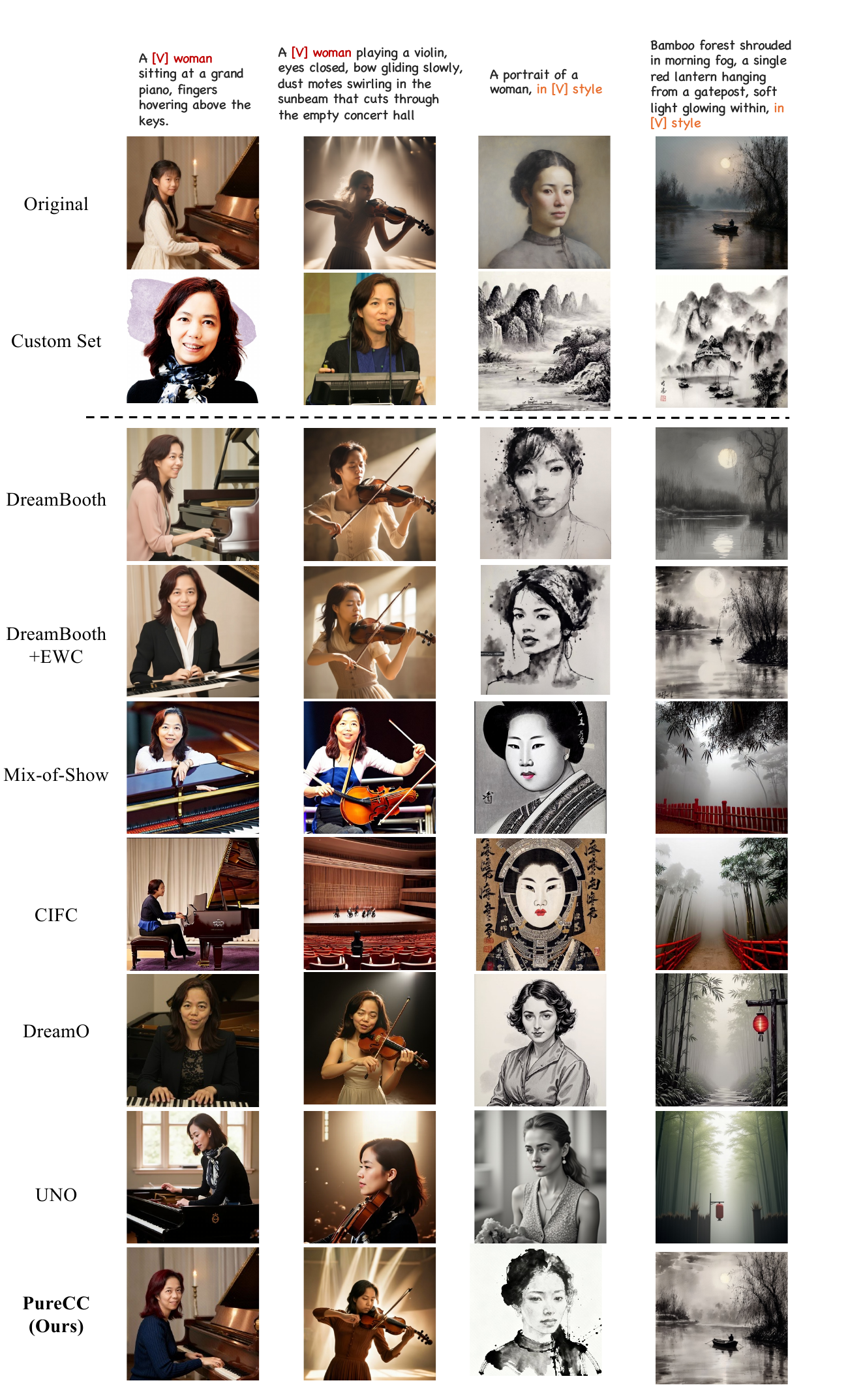}
  \caption{\textbf{More qualitative evaluation results}}
  \label{fig:sota1}
\end{figure*}

\begin{figure*}[!t]
  \centering
  \includegraphics[width=0.8\linewidth]{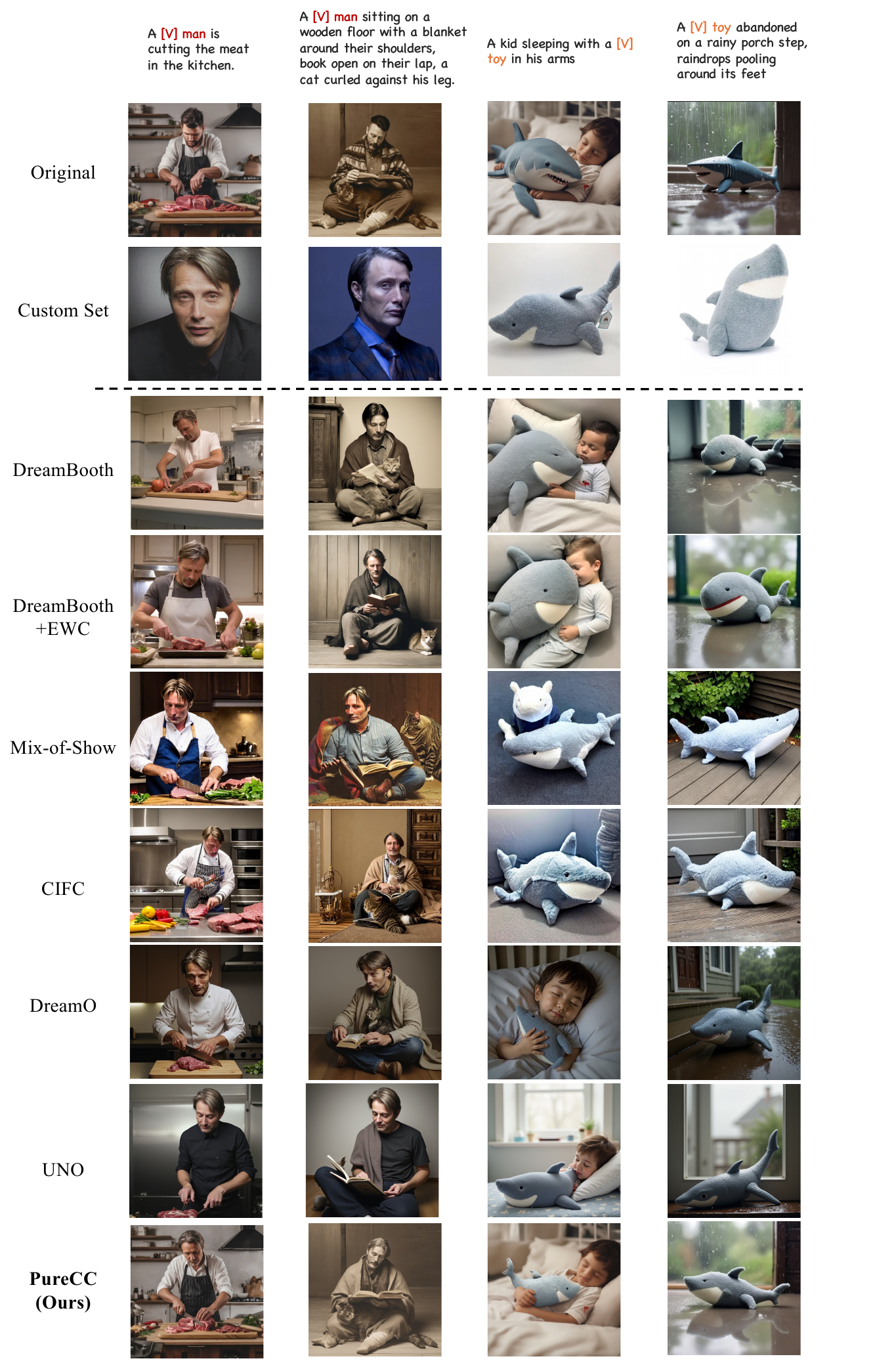}
  \caption{\textbf{More qualitative evaluation results}}
  \label{fig:sota2}
\end{figure*}

We provided more qualitative evaluation results in Fig.~\ref{fig:sota1} and Fig.~\ref{fig:sota2}.

% For a fair comparison, Tuning-based baselines such as DreamBooth~\cite{ruiz2023dreambooth}, B-LoRA~\cite{frenkel2024implicit}, LoRA-S~\cite{zhong2024multi}, and Mix-of-Show~\cite{gu2023mix} use the same pretrained backbone SD 3.5-M.
% And Tuning-free baselines such as DreamO~\cite{mou2025dreamo} and UNO~\cite{wu2025less}, we selected one image of the custom sets as the reference image to achieve customization.
% 对于风格化的样本，我们在base text前加入“Generate a same style image.”. 例如，base text:"A magical forest with glowing mushrooms and fairies.” $\rightarrow$ "Generate a same style image. A magical forest with glowing mushrooms and fairies.”
% \section{Quantitative Evaluation}
% 数据集
% baseline setting
% 

\clearpage

\section{Computational Cost}
% 计算开销
% 相对于先前的方法，我们的方法引入了一个额外的训练阶段，并且在Pure Learning（Stage-2）阶段引入一个额外的的模型，不可避免带来更多的训练时间和显存消耗。为此，我们在这里emphasize that 1）虽然需要额外的训练时间，但对于单个属性，我们完整的一次训练只需要0.33 A100 hour，仍然是高校的。2）虽然pure learing双分支需要占用额外的显存，但实际上在训练时只需要加载在一个主网络DiT，以及$\bm{v}^{\theta_1}_{t}$和$$\bm{v}^{\theta_2}_{t}$$各自对应的LoRA。因此显存消耗不会有明显增加。
Compared with previous approaches, our method introduces an additional training stage and employs an additional model branch in the Pure Learning (Stage-2) phase, which inevitably increases training time and GPU memory usage. To clarify, as shown in Tab.~\ref{tab:cost}, we emphasize that:
1) Although an extra training stage is required, completely training a single personalized concept using PureCC only takes 0.33 A100 hours, which remains highly efficient.
2) While the dual-branch design in Pure Learning indeed requires additional memory, in practice we only need to load one main network (DiT) along with the LoRA modules corresponding to ${\bm{v}}_{t}^{\theta_1}(\cdot)$ and ${\bm{v}}_{t}^{\theta_2}(\cdot)$. Therefore, the overall GPU memory consumption does not increase significantly.

\begin{table}[!h]
\centering
\caption{\textbf{Computation Time and Memory Usage of Training under BFloat16 datatype.}}
\label{tab:cost}
\resizebox{\linewidth}{!}{
\begin{tabular}{lcc}
\toprule
& Representation Extrator (stage-1) & Pure Learning (stage-2)\\
\cmidrule{2-3}
\multicolumn{1}{c}{\multirow{1}{*}{\textbf{Memory Cost}}}  &28GB & 30GB\\
\midrule
\textbf{Training Time} & 0.13 (A100 Hour) & 0.2 (A100 Hour) \\ %(12/18 * 0.2)
 \bottomrule
\end{tabular}
}
\end{table}

Tab.~\ref{tab:Computational_efficiency} compares the computational cost of training and inference with the baselines, showing that our training remains efficient.
During inference, we only need to use the single model ${\bm{v}}_{t}^{\theta_2}(\cdot)$, thus incurring no additional overhead over the baselines.

% \vspace{-1em}
\begin{table}[ht]
\centering
\caption{\textbf{Comparisons on a single NVIDIA A100 GPU}}
\label{tab:Computational_efficiency}
% \vspace{-1em}
\resizebox{1\linewidth}{!}
{
\begin{tabular}{c|cccc|c}
\hline
                  & DreamBooth & LoRA & Mix-of-Show & CIFC & \textbf{PureCC (Ours)} \\ \hline
\textbf{Training} 
& 58.0G / 0.25h 
& 28.0G / 0.13h 
& 32.0G / 0.23h 
& 31.0G / 0.28h 
& \textbf{30.0G / 0.33h} \\ \hline
\textbf{Inference} 
& 17.4G / 4.46s 
& 17.8G / 4.72s 
& 18.4G / 5.80s 
& 18.0G / 5.48s 
& \textbf{17.8G / 4.72s} \\ \hline
\end{tabular}
}
\end{table}
\vspace{-1em}

% Thus, with virtually no loss in generation quality, we adopt ${\bm{v}}_{t}^{\theta_2}(x_t|y_{base})$ as the Original Conditional Prediction throughout our Pure Learning pipeline 因为更加节约计算资源.

% the predictions produced by ${\bm{v}}_{t}^{\theta_2}(x_t|y_{base})$ closely match those obtained from the frozen pretrained model ${\bm{v}}_{t}^{\theta_3}(x_t|y_{base})$.

\section{Analysis of Hyperparameter $\eta$ in $\mathcal{L}_{PCC}$}
\begin{figure*}[!t]
  \centering
  \includegraphics[width=\linewidth]{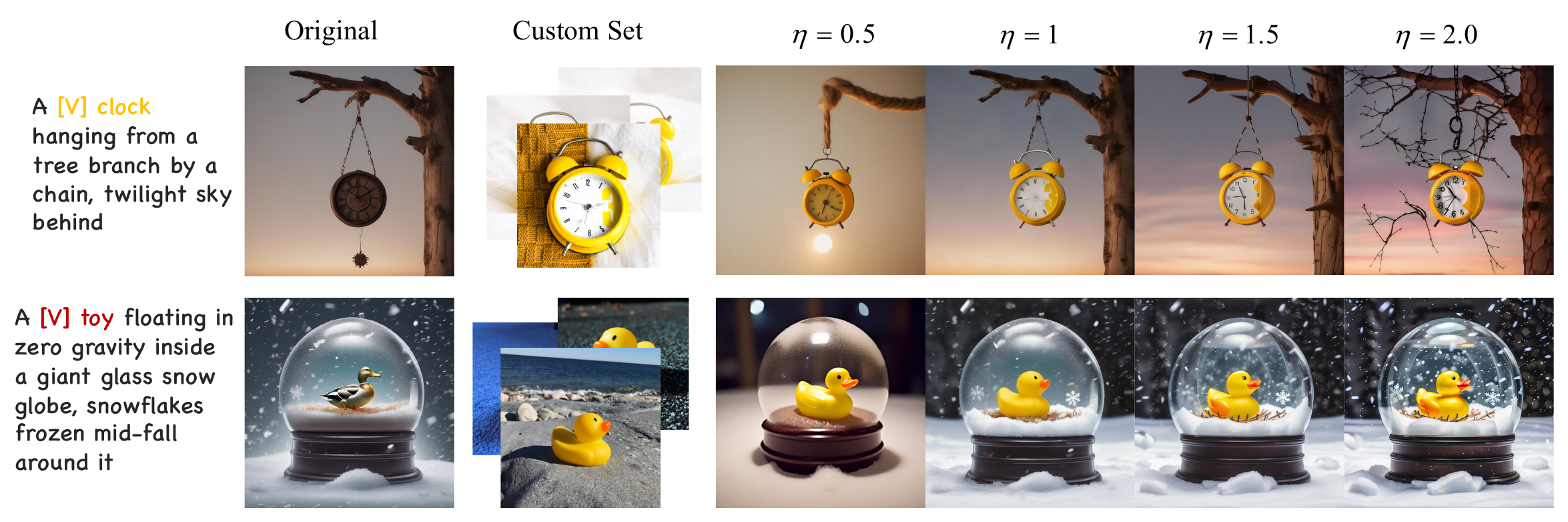}
  \caption{\textbf{Qualitative Analysis of the $\eta$.}}
  \label{fig:chaocan}
\end{figure*}

% 由于我们的Pure Learning 的loss $\mathcal{L}_{PCC}$ 引入了$\eta$ 调控纯净概念引导loss $\mathcal{L}_{PureCC}$，为了探寻$\eta$的敏感性，我们进行定性，如Fig.~\ref{fig:chaocan}和定量实验，如\ref{tab:abla2}所示，过高的$eta$对概念的过度注入会影响画面的保真度，而过小的$\eta$会导致模型受到$\mathcal{L}_{custom}$，破坏原模型行为。
Since our Pure Concept Customization loss $\mathcal{L}_{PCC}$ introduces a weighting parameter $\eta$ to modulate the pure learning loss $\mathcal{L}_{PureCC}$, we further analyze the sensitivity of $\eta$.
As shown in the qualitative results in Fig.~\ref{fig:chaocan} and the quantitative comparison in Tab.~\ref{tab:abla2}, an excessively large $\eta$ leads to over-injection of the target concept, harming visual fidelity, whereas an overly small $\eta$ causes the model to be dominated by $\mathcal{L}_{CC}$, resulting in the degradation of the original model’s behavior and capabilities.

\definecolor{tb-blue}{RGB}{196, 228, 252}
\begin{table*}[ht]
\centering
\caption{
\textbf{Quantitative of the $\eta$. }
}
% \vspace{-1em}
\label{tab:abla2}
\resizebox{0.66\textwidth}{!}
{
\begin{tabular}{lcccc}
\toprule
\multicolumn{1}{c}{\multirow{2}{*}{\textbf{Strategy}}}& \multicolumn{2}{c}{\textbf{Instance}} & \multicolumn{2}{c}{\textbf{Style}}\\
\cmidrule(r){2-3} \cmidrule(r){4-5}
  & $\Delta$CLIP-T (base) ($\uparrow$)  & CLIP-I (target) ($\uparrow$) & $\Delta$CLIP-T (base) ($\uparrow$) & CSD ($\uparrow$) \\
\midrule
$\eta = 0.5$ & -1.32 & 0.55 & -2.02 & 0.49  \\
\rowcolor{tb-blue!20}
$\eta = 1.0$ & -0.31 & 0.81 &-0.26 & 0.63  \\
$\eta = 1.5$ & -0.43 & 0.80 &-0.26 & 0.60\\
$\eta = 2.0$ & -0.76 & 0.67 &-0.41 & 0.39\\
\bottomrule
\end{tabular}
}
\end{table*}

% \section{Evaluation of Segmentation-Based Alternatives to PureCC}
% Prior methods such as PuLID and DreamO attempt to isolate the target subject through segmentation-based preprocessing, aiming to prevent the model from learning spurious correlations between the foreground object and background noise. While such strategies can reduce background entanglement at the data level, they do not fundamentally guarantee pure concept customization.
% To verify this, we replace our PureCC loss with a segmentation-driven preprocessing pipeline and evaluate whether this alternative can achieve the goal of Pure Concept Customization. 
% 先前有的方法通过分割出需要学习的主体，如PuLID, DreamO，这些方法从数据预处理上预防了模型学习到嘈杂的background与主体的耦合。但这样的方法并不能xxx，为此，我们尝试使用该策略替代CVDA loss验证是否能够实现Pure Concept Customization 这个任务。

\section{Analysis of the Original Conditional Prediction $\bm{v}_t^{original}$}
\begin{figure*}[!h]
  \centering
  \includegraphics[width=0.85\linewidth]{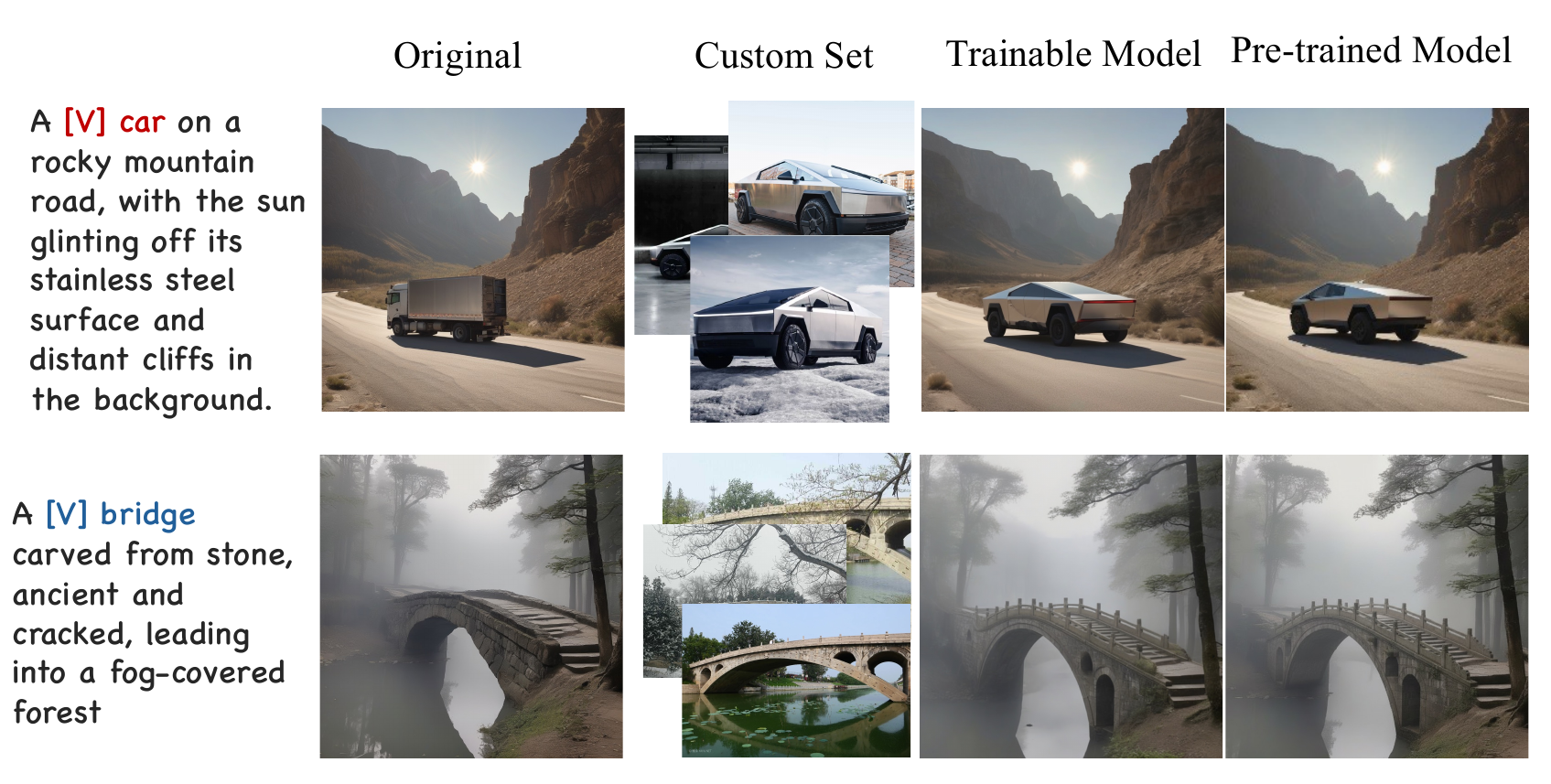}
  \caption{\textbf{Results compared} to which Original Conditional Prediction is provided by the Trainable model ${\bm{v}}_{t}^{\theta_2}(x_t|y_{base})$ or another Frozen pre-trained model ${\bm{v}}_{t}^{\theta_3}(x_t|y_{base})$ .}
  \label{fig:original}
\end{figure*}

% 再引入一个frozen模型 作为原始目标
% 正文中$\bm{v}_t^{original} = \bm{v}_t^{\theta_2}(x_t\mid y_{base})$，来自于pure customization model的输出。
% 一种更为直观的策略时，令$\bm{v}_t^{original} = \bm{v}_t^{\theta}(x_t\mid y_{base})$，通过冻结的原模型来给出 Original Conditional Prediction.
% 然而这种行为会为阶段二的pure learning再次引入一个模型，造成计算时间上的进一步消耗。
% 因此，我们验证了$\bm{v}_t^{\theta_2}(x_t\mid y_{base})$能否代表原模型对base文本的理解。如图xx所示，使用$\bm{v}_t^{\theta_2}(x_t\mid y_{base})$的效果几乎与$\bm{v}_t^{\theta}(x_t\mid y_{base})$的一致。因此在几乎不影响最后生成效果的情况下，我们使用$\bm{v}_t^{\theta_2}(x_t\mid y_{base})$输出Original Conditional Prediction.
In the main paper, we employ $\bm{v}_t^{original} = {\bm{v}}_{t}^{\theta_2}(x_t|y_{base})$, treating the output of the trainable model as the original conditional prediction.
A more effective and intuitive strategy would be to use $\bm{v}_t^{original} = {\bm{v}}_{t}^{\theta_3}(x_t|y_{base})$, i.e., obtain the original conditional prediction directly from another frozen pre-trained model.
However, this approach would introduce an additional model into the stage of Pure Learning, resulting in a non-negligible increase in computational cost.
Therefore, we evaluate whether ${\bm{v}}_{t}^{\theta_2}(x_t|y_{base})$ can reliably approximate the performance of the original model.
As shown in Fig.~\ref{fig:original}, the effect of pure learning achieved when the Original Conditional Prediction uses ${\bm{v}}_{t}^{\theta_2}(x_t|y_{base})$ from the trainable model closely matches the effect of introducing ${\bm{v}}_{t}^{\theta_3}(x_t|y_{base})$ from an additional frozen pre-trained model.
Thus, with virtually no loss in generation quality, we adopt ${\bm{v}}_{t}^{\theta_2}(x_t|y_{base})$ 
as the Original Conditional Prediction throughout our Pure Learning pipeline because it is more computationally efficient.

\section{Ablation Study Details}
We provide a detailed explanation of our ablation study here.
To demonstrate the effectiveness of our proposed novel learning objective, we designed an ablation experiment for $\mathcal{L}_{PureCC}$. This experiment involves fine-tuning using only the traditional $\mathcal{L}_{CC}$ for concept customization and comparing it with fine-tuning using our complete learning objective $\mathcal{L}_{PCC}$ ($\mathcal{L}_{CC} + \mathcal{L}_{PureCC}$).
Furthermore, to demonstrate the effectiveness of our training strategy—specifically, first obtaining a target concept representation extractor through fine-tuning on the custom set and then keeping it frozen during the pure learning stage to stably provide purified target concept representations. We designed a comparative experiment: the `Merged Learning Stage' experiment.
In the `Merged Learning Stage' experimental setup, we do not pre-fine-tune the representation extractor on the custom set. Instead, we directly use a pre-trained flow model as the representation extractor $\bm{v}^{\theta_1}_{t}$. 
During the pure learning stage, it remains trainable, allowing it to simultaneously learn the target concept and provide target concept representations for the pure learning branch $\bm{v}^{\theta_2}_{t}$.
By comparing this jointly conducted approach with our proposed method, we can demonstrate the effectiveness of dividing our training into two distinct stages.

\begin{figure*}[!h]
  \centering
  \includegraphics[width=0.88\linewidth]{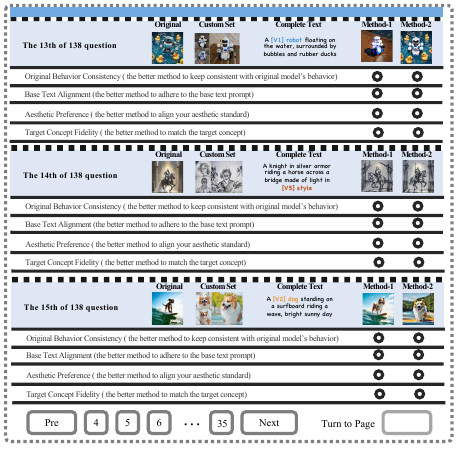}
  \caption{\textbf{The investigation page in user study.}}
  \label{fig:user_page}
\end{figure*}

\begin{table*}[!h]
\centering
% \vspace{2em}
\caption{
\textbf{The results of User Study.} (PureCC vs SOTAs)
}
% \vspace{-1em}
\label{tab:user}
\renewcommand{\arraystretch}{1.78} % 调整行间距
\resizebox{\textwidth}{!}
{
\begin{tabular}{l|cc|cc|cc|cc}
\toprule
\multicolumn{1}{c|}{}         & \large \textbf{PureCC (Ours)} & \large DreamBooth & \large \textbf{PureCC (Ours)} & \large DreamBooth+EWC & \large \textbf{PureCC (Ours)} & \large Mix-of-Show & \large \textbf{PureCC (Ours)} & \large CIFC   \\ \midrule
\large \textbf{Original Behavior Consistency} & \large 98.5\%       & \large 1.5\%      & \large 96.9\%       & \large 3.1\%          & \large 91.5\%       & \large 8.5\%       & \large 94.6\%       & \large 5.4\%  \\
\large \textbf{Base Text Alignment}           & \large 66.2\%       & \large 33.8\%     & \large 62.3\%       & \large 37.7\%         & \large 55.4\%       & \large 44.6\%      & \large 58.5\%       & \large 41.5\% \\
\large \textbf{Aesthetic Preference}          & \large 71.9\%       & \large 28.1\%     & \large 73.5\%       & \large 26.5\%         & \large 64.3\%       & \large 35.7\%      & \large 56.4\%       & \large 43.6\% \\
\large \textbf{Target Attribute Fidelity}     & \large 67.7\%       & \large 32.3\%     & \large 75.4\%       & \large 24.6\%         & \large 52.3\%       & \large 47.7\%      & \large 54.6\%       & \large 45.4\% \\ \bottomrule
\end{tabular}
}
\end{table*}

\section{User Study}
Besides qualitative and quantitative comparisons, to thoroughly evaluate our method, we carried out a user study to determine whether our method is preferred by humans for pure concept customization. 
We engaged 42 participants from diverse social backgrounds, with each test session lasting approximately 30 minutes. 
During the investigation, as illustrated in Fig.~\ref{fig:user_page}, participants conducted pairwise comparisons between our method and competing approaches across four dimensions: (1) Original Behavior Consistency, (2) Base Text Alignment, (3) Aesthetic Preference~\cite{liao2025humanaesexpert}, and (4) Target Concept Fidelity.
For ``Original Behavior Consistency," users were asked to select which of the two images better maintained consistency with the original model’s outputs, disregarding the insertion of personalized concepts.
For ``Base Text Alignment," users were tasked with choosing which of the two images more accurately align with the base text's description.
For ``Aesthetic Preference," users determined which image better matched their aesthetic preferences, taking into account factors such as visual quality and the absence of artifacts or distortions.
For ``Target Attribute Fidelity," users assessed which image more accurately generated visual content that resembled the target concepts in the custom set.
The results, as shown in Tab.~\ref{tab:user}, demonstrate that our method significantly improves the preservation of the original model's behavior and capabilities while also achieving customization effects for the target concept that are comparable to those of existing methods focused on fidelity to personalized concepts.
This underscores our strong capability to maintain the integrity of the original model while seamlessly adapting to new concepts.
This comprehensive evaluation framework guarantees a thorough and objective assessment of our method’s performance compared to existing approaches.


\end{document}